\documentclass[letterpaper]{article} 
\usepackage{aaai23}  
\usepackage{times}  
\usepackage{helvet}  
\usepackage{courier}  
\usepackage[hyphens]{url}  
\usepackage{graphicx} 
\usepackage{natbib}  
\usepackage{caption} 
\usepackage{algorithm}
\usepackage[noend]{algpseudocode}
\usepackage{amsmath}
\usepackage{amsthm}
\usepackage{amssymb,amsfonts}
\usepackage{xspace}
\usepackage{enumitem}
\usepackage{multicol}
\usepackage{multirow}
\usepackage{url}
\usepackage{soul}

\usepackage{nicematrix}
\usepackage{tikz}

\newtheorem{theorem}{Theorem}

\newtheorem{condition}{Condition} 

\newtheorem{remark}{Remark}
\newtheorem{definition}{Definition}
\newtheorem{claim}{Claim}
\newtheorem*{lemma*}{Lemma}
\newtheorem*{theorem*}{Theorem}
\newtheorem{example}{Example}
\newtheorem{lemma}{Lemma}

\newcommand{\lovasz}{{Lov\'asz}\xspace}
\newcommand{\nls}{\textsc{Nelson}\xspace}

\newcommand{\var}{\mathtt{var}}
\newcommand{\p}{\mathbb{P}}
\newcommand{\e}{\mathbb{E}}

\usepackage{newfloat}
\usepackage{listings}
\DeclareCaptionStyle{ruled}{labelfont=normalfont,labelsep=colon,strut=off} 
\floatstyle{ruled}
\newfloat{listing}{tb}{lst}{}
\floatname{listing}{Listing}
\title{Learning Markov Random Fields for Combinatorial Structures via \\
 Sampling through \lovasz Local Lemma}
\author{
    Nan Jiang\equalcontrib\textsuperscript{\rm 1},
    Yi Gu\equalcontrib\textsuperscript{\rm 2},
    Yexiang Xue\textsuperscript{\rm 1}
}
\affiliations{
    \textsuperscript{\rm 1} Department of Computer Science, Purdue University, USA\\
    \textsuperscript{\rm 2} Department of Mathematics, Northwestern University, USA \\
    \{jiang631, yexiang\}@purdue.edu, Yi.Gu@u.northwestern.edu
}

\begin{document}

\maketitle

\begin{abstract}

Learning to generate complex combinatorial structures satisfying constraints will have transformative impacts in many application domains.
However, it is beyond the capabilities of existing approaches due to the highly intractable nature of the embedded probabilistic inference.
Prior works spend most of the training time learning to separate valid from invalid structures but do not learn the inductive biases of valid structures.
We develop   \underline{NE}ural \underline{L}ov\'asz \underline{S}ampler (\nls), which embeds the sampler through \lovasz Local Lemma (LLL) as a fully differentiable neural network layer.
Our \nls-CD embeds this sampler 
into the contrastive divergence learning process of Markov random fields.
\nls allows us to obtain valid samples from the current model distribution. 
Contrastive divergence is then applied to separate these samples from those in the training set. 
\nls is implemented as a fully differentiable neural net, taking advantage of the parallelism of GPUs.
Experimental results on several real-world domains reveal that \nls learns to generate 100\% valid structures, while baselines either time out or cannot ensure validity. 
\nls also outperforms other approaches in running time, log-likelihood, and MAP scores.
\end{abstract}
\section{Introduction}
In recent years, tremendous progress has been made in generative modeling~\cite{DBLP:journals/neco/Hinton02,Tsochantaridis2005StructualSVM,goodfellow2014generative,kingma2013auto,Germain2015MADEMA,Larochelle2011Autoregressive,Oord2016PixelRN,arjovsky17WGAN,song2019generative,song21scoredifussion,murphy1999loopy,yedidia2001generalized,wainwright2006relaxation}. 

Learning a generative model involves increasing the divergence in  likelihood scores  between the structures in the training set and those structures sampled from the current generative model distribution. 
While current approaches have achieved successes in \textit{un-structured} domains such as vision or speech, their performance is degraded in the structured domain, because it is already computationally intractable to search for a valid  structure in a combinatorial space subject to constraints, not to mention sampling, which has a higher complexity.
%
%
In fact, when applied in a constrained domain, existing approaches spend most of their training time manipulating the likelihood of invalid structures, but not learning the difference between valid structures inside and outside of the training set. 
In the meantime, tremendous progress has been made in automated reasoning \cite{Braunstein2005SurveyPA,AnHaHoTi2007,chavira2006compiling,VanHentenryck1989,GOGATE2012AndOr,Sang2005}. 
Nevertheless, reasoning and learning have been growing independently for a long time. Only recently do ideas  emerge exploring the role of reasoning in learning~\cite{kusner17GrammarVAE,Jin2018JunctionTV,Dai2018SyntaxDirectedVA,Hu17ControlledTextGen,lowd2008learning,DBLP:conf/icml/DingMXX21}.

The \lovasz Local Lemma (LLL)~\cite{erdHos1973problems} is a classic gem in combinatorics, which at a high level, states that there exists a positive probability that none of a series of \textit{bad} events occur, as long as these events are \textit{mostly} independent from one another and are not too likely individually. 
Recently, \citet{DBLP:journals/jacm/MoserT10} came up with an algorithm, which samples from the probability distribution proven to exist by LLL.
\citet{DBLP:journals/jacm/GuoJL19} proved that the algorithmic-LLL is an unbiased sampler 
if those bad events satisfy the so-called ``extreme'' condition. 
The expected running time of the sampler is also shown to be polynomial. 
As one contribution of this paper, we offer proofs of the two aforementioned results using precise mathematical notations, clarifying a few descriptions not precisely defined in the original proof. 
While this line of research clearly demonstrates the potential of LLL in generative learning (generating samples that satisfy all hard constraints), it is not clear how to embed LLL-based samplers into learning and no empirical studies have been performed to evaluate LLL-based samplers in machine learning.

In this paper, we develop   \underline{NE}ural \underline{L}ov\'asz \underline{S}ampler (\nls), which implements the LLL-based sampler as a fully differentiable neural network. 
Our \nls-CD embeds \nls 
into the contrastive divergence learning process of Markov Random Fields (MRFs). 
%
%
Embedding LLL-based sampler allows the contrastive learning algorithm to focus on learning the difference  between the training data and the valid structures drawn from the current model distribution. Baseline approaches, on the other hand, spend most of their training time learning to generate valid structures. 
In addition, \nls is fully differentiable, hence allowing for efficient learning harnessing the parallelism of GPUs.

Related to our \nls are neural-based approaches to solve combinatorial optimization problems~\cite{DBLP:conf/iclr/SelsamLBLMD19,DBLP:conf/icml/DuanVPRM22,DBLP:conf/nips/LiCK18}. 
Machine learning is also used to discover better heuristics~\cite{DBLP:conf/nips/YolcuP19,DBLP:conf/nips/ChenT19}.
Reinforcement learning \cite{DBLP:conf/nips/KaraliasL20,DBLP:journals/eor/BengioLP21} as well as approaches integrating search with neural nets \cite{DBLP:conf/aaai/MandiDSG20} are found to be effective in solving combinatorial optimization problems as well.
Regarding probabilistic inference, there are a rich line of research  on  MCMC-type sampling~\cite{neal1993probabilistic,DBLP:journals/pami/DagumC93,pmlr-v84-ge18b} and various versions of belief propagation~\cite{murphy1999loopy,DBLP:journals/jmlr/IhlerFW05,DBLP:journals/corr/abs-2011-02303,pmlr-v138-fan20a}. 
SampleSearch~\cite{DBLP:journals/ai/GogateD11} integrates importance sampling with  constraint-driven search.
Probabilistic inference based on hashing and randomization obtains probabilistic guarantees for marginal queries and sampling via querying optimization oracles subject to randomized constraints \cite{Gomes2006Sampling,Ermon13Wish,DBLP:conf/sat/AchlioptasT17,Chakraborty2013scalable}.  

We experiment \nls-CD on learning preferences towards (i) random $K$-satisfiability solutions (ii) sink-free orientations of un-directed graphs and (iii) vehicle delivery routes. 
In all these applications, \nls-CD 
(i) has the fastest training time due to seamless integration into the learning framework (shown in Tables~\ref{tab:sampler}(a),~\ref{tab:sink-free}(a)).  (ii) \nls generates samples 100\% satisfying constraints 
 (shown in Tables~\ref{tab:sampler}(b),~\ref{tab:sink-free}(b)), which facilitates effective contrastive divergence learning. Other baselines either cannot satisfy constraints or time out. (iii) The fast and valid sample generation allows \nls to obtain the best  learning performance (shown in Table~\ref{tab:sampler}(c),~\ref{tab:learn-sat}(a,b),~\ref{tab:sink-free}(c,d)).

Our contributions can be summarized as follows: 
\textbf{(a)} We present \nls-CD, a contrastive divergence learning algorithm for constrained MRFs driven by sampling through the \lovasz Local Lemma (LLL). \textbf{(b)} Our LLL-based sampler (\nls) is implemented as a fully differentiable multi-layer neural net, allowing for end-to-end training on GPUs. \textbf{(c)} We offer a mathematically sound proof of the sample distribution and the expected running time of the \nls algorithm.
\textbf{(d)} Experimental results reveal the effectiveness of \nls in  (i) learning models  with high likelihoods (ii) generating samples 100\% satisfying constraints and (iii) having high efficiency in training\footnote{Code is at: \url{https://github.com/jiangnanhugo/nelson-cd}.\\ {Please refer to the Appendix in the extended version~\cite{nanyiye2022} for the whole proof and the experimental settings.}}.

\section{Preliminaries} \label{sec:prelim}
\subsubsection{Markov Random Fields (MRF)} represent a Boltzmann distribution of the discrete variables  $X=\{X_i\}_{i=1}^n$ over a Boolean hypercube 
 $\mathcal{X}=\{0, 1\}^n$. For $x\in\mathcal{X}$, we have:
\begin{equation} \label{eq:mrf}
P_{\theta}(X=x)=\frac{\exp \left(\phi_\theta(x)\right)}{Z(\theta)}=\frac{\exp \left(\sum_{j=1}^{m} \phi_{\theta,j}(x_{j})\right)}{Z(\theta)}.
\end{equation}
Here,
$Z(\theta)=\sum_{x'\in \mathcal{X}} \exp\left(\phi_\theta(x')\right)$ is the partition function that normalizes the total probability to $1$. {The potential function is} $\phi_\theta(x)=\sum_{j=1}^m \phi_{\theta,j}(x_{j})$. 
Each $\phi_{\theta,j}$ is a \textit{factor potential}, which maps a value assignment over a subset of variables $X_j \subseteq X$ to a real number. 
We use upper case letters, such as $X_j$ to represent (a set of) random variables, and use lower case letters, such as $x_j$, to represent its value assignment. 
We also use $\var(\phi_{\theta, j})$ to represent the domain of $\phi_{\theta, j}$, \textit{i.e.}, $\var(\phi_{\theta, j}) = X_j$.
$\theta$ are the parameters to learn.

\subsubsection{Constrained MRF} is the MRF model subject to
a set of hard constraints $\mathcal{C}=\{c_j\}_{j=1}^L$.
Here, each constraint $c_j$ limits the value assignments of a subset of variables $\var(c_j) \subseteq X$.
We write $c_j(x) = 1$ if the assignment $x$  satisfies the constraint $c_j$ and $0$ otherwise. Note that $x$ is an assignment to all random variables, but $c_j$ only depends on variables  $\var(c_j)$.
We denote $C(x) = \prod_{j=1}^L c_j(x)$ as the indicator function. Clearly, $C(x) = 1$ if all constraints are satisfied and $0$ otherwise. 
The constrained MRF is:
\begin{equation} \label{eq:constr_mrf}
P_{\theta}(X=x|\mathcal{C})=\frac{\exp \left(\phi_\theta(x)\right)C(x)}{Z_\mathcal{C}(\theta)},
\end{equation}
where $Z_\mathcal{C}(\theta)={\sum}_{x'\in \mathcal{X}} \exp\left(\phi_\theta(x)\right) C(x)$ sums over only valid assignments. 

\subsubsection{Learn Constrained MRF} 
Given a data set $\mathcal{D}=\{x^k\}_{k=1}^N$, where each $x^k$ is a valid assignment that satisfies all constraints, 
learning can be achieved via maximal likelihood estimation. In other words, we find the
optimal parameters $\theta^*$ by minimizing the negative $\log$-likelihood $\ell_{\mathcal{C}}(\theta)$:
\begin{equation}  \label{eq:log-likelihood} 
\begin{aligned}
\ell_{\mathcal{C}}(\theta)&=-\frac{1}{N}\sum_{k=1}^N\log P_{\theta}(X=x^k|\mathcal{C})\\
&=-\frac{1}{N}\sum_{k=1}^N\phi_{\theta}(x^k)+\log Z_{\mathcal{C}}(\theta).
\end{aligned}
\end{equation}
The parameters $\theta$ can be trained using gradient descent: $\theta^{t+1}= \theta^{t}-\eta\nabla\ell_{\mathcal{C}}(\theta)$, where $\eta$ is the learning rate. Let $\nabla\ell_{\mathcal{C}}(\theta)$ denotes the gradient of the objective $\ell_{\mathcal{C}}(\theta)$, that is calculated as:
\begin{equation}
\begin{aligned}\label{eq:gradient}
\nabla\ell_{\mathcal{C}}&(\theta)=-\frac{1}{N}\sum_{k=1}^N\nabla \phi_{\theta}(x^k)+ \nabla\log Z_{\mathcal{C}}(\theta)\\
&=-\mathbb{E}_{{x}\sim \mathcal{D}}\left(\nabla\phi_{\theta}({x})\right)+\mathbb{E}_{\tilde{x}\sim P_{\theta}(x|\mathcal{C})}\left(\nabla\phi_{\theta}(\tilde{x})\right).
\end{aligned}
\end{equation}
The first term is the expectation over all data in training set $\mathcal{D}$. 
During training, this is approximated using a mini-batch of data randomly drawn from the training set $\mathcal{D}$. 
The second term is the expectation  over the current model distribution $P_\theta(X=x|\mathcal{C})$ (detailed in Appendix C.2). 
Because learning is achieved following the directions given by the divergence of two expectations, this type of learning is commonly known as contrastive divergence (CD)~\cite{DBLP:journals/neco/Hinton02}. 
Estimating the second expectation is the bottleneck of training because it is computationally intractable to sample from this distribution subject to combinatorial constraints. 
Our approach, \nls, leverages the sampling through \lovasz Local Lemma to approximate the second term.

\subsubsection{Factor Potential in Single Variable Form}
Our method requires each factor potential $\phi_{\theta,j}(x_j)$ in Eq.~\eqref{eq:mrf} to involve only one variable. 
This is NOT an issue as \textit{all constrained MRF models can be re-written in single variable form} by introducing additional variables and constraints. 
Our transformation follows the idea in~\citet{Sang2005}.
We illustrate the idea by transforming one-factor potential $\phi_{\theta,j}(x_j)$ into the single variable form. 
First, notice all functions including $\phi_{\theta,j}(x_j)$ over a Boolean hypercube $\{0,1\}^n$ have a (unique) discrete Fourier expansion:
\begin{equation}
    \phi_{\theta,j}(x_j) = \sum_{S\in [{\var(\phi_{\theta,j})}]} \hat{\phi}_{\theta,j,S}~ \chi_S(x).
\end{equation}
Here $\chi_S(x)=\prod_{X_i \in S} X_i$ is the basis function and $\hat{\phi}_{\theta,j,S}$ are Fourier coefficients.
$[{\var(\phi_{\theta,j})}]$ denotes the power set of ${\var(\phi_{\theta,j})}$. For example, if $\var(\phi_{\theta,j}) = \{X_1$, $X_2\}$, then  $[{\var(\phi_{\theta,j})}]=\{\emptyset, \{X_1\}, \{X_2\}, \{X_1, X_2\}\}$. 
See \citet{Mansour1994LearningBF} for details of Fourier transformation. 
To transform $\phi_{\theta,j}(x_j)$ into single variable form, we introduce a new Boolean variable $\hat{\chi}_S$ for every $\chi_S(x)$.
Because $\hat{\chi}_S$ and all $X_i$'s are Boolean, we can use combinatorial constraints to guarantee $\hat{\chi}_S=\prod_{X_i \in S} X_i$.
These constraints are incorporated into $\mathcal{C}$. 
Afterward, $\phi_{\theta,j}(x_j)$ is represented as the sum of several single-variable factors. 
Notice this transformation is only possible when the MRF is subject to constraints. 
We offer a detailed example in Appendix C.1 for further explanation.
Equipped with this transformation, we assume all $\phi_{\theta,j}(x_j)$ are single variable factors for the rest of the paper.
 
\subsubsection{Extreme Condition}  The set of constraints $\mathcal{C}$ is called ``extremal'' if
no variable assignment  violates two constraints sharing variables, according to~\citet{DBLP:journals/jacm/GuoJL19}.
\begin{condition}
\label{cond:extreme}
    A set of constraints $\mathcal{C}$ is called extremal if and only if for each pair of constraints $c_i ,c_j \in \mathcal{C}$, (i) either their domain variables do not intersect, i.e., ${\var(c_i)} \cap {\var(c_j)} = \emptyset$.  (ii) or for all $x\in \mathcal{X}$, $c_i(x)=1$ or $c_j(x)=1$.
\end{condition}

\section{Sampling Through Lov\'asz Local Lemma}  
Lov\'asz Local Lemma (LLL)~\citep{erdHos1973problems} is a fundamental method in combinatorics to show the existence of a valid instance that avoids all the bad events, if the occurrences of these events are ``mostly'' independent and are not very likely to happen individually. 
Since the occurrence of a bad event is equivalent to the violation of a constraint, we can use the LLL-based sampler to sample from the space of constrained MRFs. 
To illustrate the idea of LLL-based sampling, we assume the constrained MRF model is given in the single variable form (as discussed in the previous section):
\begin{equation}  \label{eq:constr_mrf_single}
\begin{aligned}
P_{\theta}(X=x|\mathcal{C})&=\frac{\exp \left(\sum_{i=1}^n \theta_i x_{i}\right)C(x)}{Z_\mathcal{C}(\theta)},
\end{aligned}
\end{equation}
where $Z_\mathcal{C}(\theta)=\sum_{x'\in \mathcal{X}} \exp\left(\sum_{i=1}^n \theta_i x_{i}\right) C(x)$.

As shown in Algorithm~\ref{alg:lll-sampler}, the LLL-based sampler~\citep{DBLP:journals/jacm/GuoJL19} takes the random variables $X=\{X_{i}\}_{i=1}^n$, the parameters of constrained MRF $\theta$, and constraints $\mathcal{C}=\{c_j\}_{j=1}^L$ that satisfy  Condition~\ref{cond:extreme} as the inputs.
In Line 1 of Algorithm~\ref{alg:lll-sampler}, the sampler gives an initial random assignment of each variable following its marginal probability: $x_i\sim \frac{\exp(\theta_i x_i)}{\sum_{x_i\in \{0,1\}}\exp(\theta_i x_i)}$, for $1\le i\le n$. 
Here we mean that $x_i$ is chosen with probability mass $\frac{\exp(\theta_i x_i)}{\sum_{x_i\in \{0,1\}}\exp(\theta_i x_i)}$.
Line 2 of Algorithm~\ref{alg:lll-sampler} checks if the current assignment satisfies all  constraints in $\mathcal{C}$.
If so, the algorithm  terminates. Otherwise, the algorithm finds the set of violated constraints $S=\{c_j|c_j(x)=0,c_j \in\mathcal{C}\}$ and re-samples related variables  $X_k\in\var(S)$ using the same marginal probability, \textit{i.e.}, $x_k\sim \frac{\exp(\theta_k x_k)}{\sum_{x_k\in \{0,1\}}\exp(\theta_kx_k)}$. Here $\var(S)=\cup_{c_j\in S}~\var(c_j)$.
The algorithm repeatedly samples all those random variables violating constraints until all  the constraints are satisfied.

\begin{algorithm}[t] 
	\caption{Sampling Through Lov\'asz Local Lemma.}\label{alg:lll-sampler}
	\begin{algorithmic}[1]
	\Require{
	 Random variables $X=\{X_{i}\}_{i=1}^n$; Constraints $\mathcal{C}=\{c_j\}_{j=1}^L$; Parameters of the constrained MRF $\theta$.}
	\State $x_i\sim\frac{\exp(\theta_i x_i)}{\sum_{x_i\in \{0,1\}}\exp(\theta_ix_i)}$, for $1\le i\le n$. \Comment{initialize}
	\While{$C(x)=0$}
			\State Find all violated constraints $S\subseteq \mathcal{C}$ in $x$.
			\State {\small $x_k{\footnotesize \sim}\frac{\exp(\theta_k x_k)}{\underset{{x_k\in \{0,1\}}}{\sum}\exp(\theta_k x_k)}, \text{for } x_k \in \var(S)$.}\Comment{resample}
	\EndWhile
	\Return  A valid sample $x$ drawn from $P_{\theta}(X=x|\mathcal{C})$.
	\end{algorithmic}
\end{algorithm}

Under Condition~\ref{cond:extreme}, Algorithm~\ref{alg:lll-sampler} guarantees each sample is from the  constrained MRFs' distribution $P_\theta(X=x | \mathcal{C})$ (in Theorem~\ref{th:product-dist}). 
In Appendix A, we present the detailed proof and clarify the difference to the original descriptive proof~\cite{DBLP:journals/jacm/GuoJL19}.
\begin{theorem}[Probability Distribution] \label{th:product-dist}
Given  random variables $X=\{X_{i}\}_{i=1}^n$, constraints $\mathcal{C}=\{c_j\}_{j=1}^L$ that satisfy Condition~\ref{cond:extreme}, and the parameters of the constrained MRF in the single variable form  $\theta$. Upon termination, Algorithm~\ref{alg:lll-sampler}  outputs an assignment $x$ that is randomly drawn from the constrained MRF distribution: $x\sim P_\theta(X=x | \mathcal{C})$. 

\begin{proof}[Sketch of Proof] We first show that in the last round, the probability of obtaining two possible assignments conditioning on all previous rounds in Algorithm~\ref{alg:lll-sampler} has the same ratio as the probability of those two assignments under distribution $P_\theta(X=x | \mathcal{C})$. 
Then we show when {Algorithm~\ref{alg:lll-sampler}} ends, the set of all possible outputs is equal to the domain of non-zero probabilities of  $P_\theta(X=x | \mathcal{C})$.
Thus we conclude the execution of Algorithm~\ref{alg:lll-sampler} produces a sample from $P_{\theta}(X=x|\mathcal{C})$ because of the identical domain and the match of probability ratios of any two {valid} assignments. 
\end{proof}
\end{theorem}

The expected running time of Algorithm~\ref{alg:lll-sampler} is determined by the number of rounds of re-sampling. 
In the uniform case that $\theta_1=\ldots=\theta_n$, the running time is linear in the size of the constraints $\mathcal{O}(L)$.
The running time for the weighted case has a  closed form. We leave the details  in Appendix B.

\section{Neural Lov\'asz Sampler}
\label{sec:methodology}
We first present the proposed \underline{Ne}ural \underline{L}ov\'asz \underline{S}ampler {(\nls)} that implements the LLL-based sampler as a  neural network, allowing us to draw multiple samples in parallel on GPUs. %
We then demonstrate how \nls is embedded in CD-based learning for constrained MRFs.

\subsection{\nls: Neural Lov\'asz Sampler} \label{sec:nls}
\subsubsection{Represent  Constraints as CNF}
\nls obtains samples from the constrained MRF model in  single variable form (Eq.~\ref{eq:constr_mrf_single}). 
To simplify notations, we denote $P_{\theta}(X_i=x_i) = \frac{\exp(\theta_i x_i)} {\sum_{x_i\in \{0,1\}} \exp(\theta_i x_i)}$.
Since our constrained MRF model is defined on the Boolean hyper-cube $\{0,1\}^n$, we assume all constraints $\{c_j\}_{j=1}^L$ are given in the Conjunctive Normal Form (CNF).
Note that all propositional logic can be reformulated in CNF format with at most a polynomial-size increase.
A formula represented in CNF is a conjunction ($\wedge$) of 
clauses. 
A clause is a disjunction ($\vee$) of literals, and a literal is either a variable or its negation ($\neg$). 
Mathematically, we use $c_j$ to denote a clause and use $l_{j,k}$ to denote a literal. In this case, a CNF formula would be:
\begin{align} \label{eq:cnf}
c_1\wedge\ldots\wedge  c_L,\quad \text{where } c_j = l_{j,1}\vee\ldots\vee l_{j,K}
\end{align}
A clause is true if and only if at least one of the literals in the clause is true. 
The whole CNF is true if all clauses are true. 

We transform each step of Algorithm \ref{alg:lll-sampler} into arithmetic operations, hence encoding it as  a multi-layer neural network. 
To do that, we first need to define a few notations:
\begin{itemize}[align=left, leftmargin=0pt, labelwidth=0pt, itemindent=!]
 \item Vector of assignment  $x^t=(x^t_1, \dots, x^t_n)$, where $x_i^t$ is the assignment of variable $X_i$ in the $t$-th round of Algorithm~\ref{alg:lll-sampler}. $x^t_i=1$ denotes  variable $X_i$ takes value $1$ (or true).
 \item Vector of marginal probabilities $P=(P_1, \ldots, P_n)$, where $P_i$ is the probability of variable $X_i$ taking value $0$ (false): $P_i = P_\theta(X_i=0) = {\exp(0)}/{(\exp(0)+\exp(\theta_i))}$.
 \item Tensor $W\in\{-1, 0, 1\}^{L\times K\times n}$ and matrix $b\in\{0,1\}^{L\times n}$, that are  used for checking constraint satisfaction: 
\begin{align}
W_{jki}&=\begin{cases}
1 &\text{if $k$-th literal of clause $c_j$ is  }  X_i,\\
-1 & \text{if $k$-th literal of clause $c_j$ is } \neg X_i, \\
0&\text{otherwise}.
\end{cases} \label{eq:weight}\\
b_{jk}&=\begin{cases}
1 &  \text{if $k$-th literal of clause $c_j$ is negated}, \\
0 & \text{otherwise}.
\end{cases}\label{eq:bias}
\end{align}
\item Matrix $V\in\{0,1\}^{L\times n}$, denoting the mapping from clauses to variables in the CNF form for constraints $\mathcal{C}$:
\begin{equation} \label{eq:mapping-matrix}
V_{ji}\mbox{=}\begin{cases}
1 &  \text{if clause $c_j$ contains a literal involving $X_i$} \\
0 & \text{otherwise}. 
\end{cases}
\end{equation}
\item Vector of resampling indicators $A^t$, where $A^t_{i}=1$ indicates  variable $X_i$ needs to be resampled at round $t$.
\end{itemize}
Given these defined variables, we represent each step of Algorithm \ref{alg:lll-sampler} using arithmetic operations as follows:

\subsubsection{Initialization} To complete line 1 of Algorithm~\ref{alg:lll-sampler}, given the marginal probability vector $P$, the first step is sampling an initial assignment of $X$,  $x^1=(x^1_1, \ldots, x^1_n)$. It is accomplished by: for $1\le i\le n$, 
\begin{equation}
 x^1_{i} = \begin{cases}
 1&\text{if } u_i> P_i,\\
 0 &\text{otherwise}. \\
 \end{cases}
\end{equation}
Here $u_i$ is sampled from the uniform distribution in $[0,1]$.

\subsubsection{Check Constraint Satisfaction} To complete line 2 of Algorithm~\ref{alg:lll-sampler}, given an assignment $x^t$ at round $t\ge 1$, tensor $W$ and matrix $b$, we compute $Z^t$ as follows:
\begin{equation}\label{eq:Zt}
Z^t =W \circledast  x^t+b,
\end{equation}
where  $\circledast$ represents a special  multiplication between tensor and vector: $(W \circledast x)_{j k}=\sum_{i=1}^n W_{j k i} x^t_i$. 
Note that $Z^t_{jk}=1$ indicates the $k$-th literal of $j$-th clause is true (takes value $1$). Hence, we compute $S^t_j$ as:
\begin{align}\label{notation:C}
S^t_j &=1-\max_{1\le k\le K} Z_{jk}, \quad \text{ for } 1\le j\le L.
\end{align}
Here $S^t_j=1$ indicates $x^t$ violates $j$-th clause.
We  check $\sum_{j=1}^LS^t_j\neq 0$ to see if any clause is violated, which corresponds to $C(x)= 0$ and is the continuation criteria of the while loop. 

\subsubsection{Extract Variables in Violated Clauses} To complete line 3 of  Algorithm~\ref{alg:lll-sampler},  we extract all the  variables that require resampling based on vector $S^t$ computed from the last step.
The vector of resampling indicator $A^t$ can be computed as:
\begin{equation}
A^t_i=\mathbf{1}\left(\sum_{j=1}^L{S_j^t} V_{ji}\ge 1\right),\quad \text{ for } 1\le i\le n
\end{equation}
where $\sum_{j=1}^L{S_j^t} V_{ji}\ge 1$ implies $X_i$ requires resampling.

\subsubsection{Resample} To complete line  4 of  Algorithm~\ref{alg:lll-sampler}, given the marginal probability vector $P$, resample indicator vector $A^t$ and assignment $x^t$, we draw a new random sample $x^{t+1}$. This can be done using this update rule: for $1\le i\le n$,
\begin{equation} \label{eq:iterative}
 x_i^{t+1}=\begin{cases}
 (1-A^t_i) x_i^t+A^t_i  &\text{if } u_i>P_i,\\
 (1-A^t_i) x_i^t &\text{otherwise}.
 \end{cases}
\end{equation}
Again, $u_i$ is drawn from the uniform distribution in $[0, 1]$. 
Drawing multiple assignments in parallel is attained by extending $x^t$ with a new dimension (See implementation in Appendix~D.1).
Example~\ref{example:matrix} show the detailed steps of \nls (See more examples in Appendix~A.5).

\begin{example}\label{example:matrix}
Assume we have random variables $X_1, X_2,X_3$ with $n=3$, Constraints $\mathcal{C}=(X_1\vee X_2)\wedge(\neg X_1 \vee X_3)$ in the CNF form with $L=2,K=2$.
Tensor $W$ is:
\begin{equation*}
\begin{aligned}
W\mbox{=}\begin{bmatrix}
w_{11}\mbox{=}[w_{111}, w_{112}, w_{113}], & w_{12}\mbox{=}[w_{121}, w_{122}, w_{123}]\\
w_{21}\mbox{=}[w_{211}, w_{212}, w_{213}], & w_{22}\mbox{=}[w_{221}, w_{222}, w_{223}] \\
\end{bmatrix},
\end{aligned}
\end{equation*}
\begin{equation*}
\begin{aligned}
 w_{11}=[1, 0, 0], w_{12}=[0, 1, 0],  w_{21}\mbox{=}[-1, 0, 0], w_{22}\mbox{=}[0 , 0, 1].
\end{aligned}
\end{equation*}
Note that $w_{111}=1$ means $X_1$ is the 1st literal in the 1st clause and $w_{211}=-1$ means $\neg X_1$ is the 1st literal in the 2nd clause. 
Matrix $b$ and the mapping matrix $V$ are:
\begin{equation*}
b=\begin{bmatrix}
0 & 0\\
1 & 0 \\
\end{bmatrix},\quad 
V=\begin{bmatrix}
1 & 1 & 0 \\
1 & 0 & 1 \\
\end{bmatrix},
\end{equation*}
$b_{21}=1$ indicates the 1st literal in the 2nd clause is negated. For the mapping matrix, $V_{11}=V_{12}=1$ implies the 1st clause contains $X_1$ and $X_2$.
For $t=1$, suppose we have an initialized assignment $x^1=[0\; 0\; 1]^\top$, meaning $X_1=X_2=0, X_3=1$. 
The intermediate results of  $Z^1,S^1,A^1$ become:
\begin{equation*}
Z^1=\begin{bmatrix}
0 & 0 \\
1 & 1 \\
\end{bmatrix},\quad 
S^1=\begin{bmatrix}
1 \\ 
0 \\
\end{bmatrix},\quad 
A^1=\begin{bmatrix}
1\\
1 \\
0 \\
\end{bmatrix},
\end{equation*}
where $S^1_{1}=1$ implies the $1$st clause is violated. $A^1_{1}=A^1_{2}=1$ denotes variables $X_1,X_2$ require resampling.
\end{example}

\begin{algorithm}[!t]
	\caption{Learn Constrained MRFs via \nls-CD.}\label{alg:main}
	\begin{algorithmic}[1]
 \Require{Dataset $\mathcal{D}$; Constraints $\mathcal{C}$;  \#Samples $m$;   Learning Iterations $T_{\max}$;  {Parameters of} Constrained MRFs $\theta$.}
	 \State  $\nls(W, b, V)\leftarrow \text{build}(X,\mathcal{C})$. \Comment{in Sec.~\ref{sec:methodology}}
	\For{$t=1$ \textbf{to} $T_{\max}$}
            \State  $\{{x}^j\}_{j=1}^m\sim \mathcal{D}$. \Comment{from data}
			\State $\{\tilde{x}^j\}_{j=1}^m\leftarrow\nls(\theta^t, m)$. \Comment{from model}
			\State  $g^t\leftarrow \frac{1}{m}\sum_{j=1}^m \nabla\phi(x^j)- \nabla\phi(\tilde{x}^j)$ \Comment{divergence} 
			\State $\theta^{t+1}\leftarrow \theta^{t}-\eta g^t$. \Comment{update parameters}
	\EndFor
	\Return  The converged MRF model $\theta^{T_{\max}}$. 
	\end{algorithmic}
\end{algorithm}

\subsection{Contrastive Divergence-based Learning}
The whole learning procedure is shown in Algorithm~\ref{alg:main}. At every learning iteration, we call \nls to draw assignments $\{\tilde{x}^j\}_{j=1}^m$ from constrained MRF's distribution $P_{\theta}(X|\mathcal{C})$.
Then we pick $m$ data points at random from the training set $\{x^j\}_{j=1}^m\sim \mathcal{D}$. The divergence $g^t$ in line 5 of Algorithm~\ref{alg:main} is an estimation of $\nabla\ell_{\mathcal{C}}(\theta)$ in Eq.~\eqref{eq:gradient}. Afterward, the MRFs' parameters are updated, according to line 6 of Algorithm~\ref{alg:main}. After $T_{\max}$ learning iterations, the algorithm outputs the constrained MRF model with parameters $\theta^{T_{\max}}$.

\section{Experiments}
We show the efficiency of the proposed \nls  on learning MRFs defined on the solutions of  three combinatorial problems.
Over all the tasks, we demonstrate that \nls outperforms baselines on learning performance, i.e., generating structures with high likelihoods and MAP@10 scores  (Table~\ref{tab:sampler}(c),~\ref{tab:learn-sat}(a,b),~\ref{tab:sink-free}(c,d)).  
\nls also generates samples which 100\% satisfy constraints (Tables~\ref{tab:sampler}(b),~\ref{tab:sink-free}(b)). 
Finally, \nls is the most efficient sampler. Baselines either time out or cannot generate valid structures (Tables~\ref{tab:sampler}(a),~\ref{tab:sink-free}(a)).  

\subsection{Experimental Settings}
\subsubsection{Baselines} We compare \nls with other contrastive divergence learning algorithms equipped with other sampling approaches. In terms of baseline samplers, we consider:
\begin{itemize}[align=left, leftmargin=0pt, labelwidth=0pt, itemindent=!]
\item Gibbs sampler~\cite{carter1994gibbs}, which is a special case of MCMC that is  widely used in training MRF models.
    \item Weighted SAT samplers, including WAPS~\cite{DBLP:conf/tacas/GuptaSRM19}, WeightGen~\cite{DBLP:conf/aaai/ChakrabortyFMSV14} and XOR sampler~\cite{DBLP:conf/nips/ErmonGSS13,DBLP:conf/uai/DingX21}.
    \item Uniform SAT samplers, including CMSGen~\cite{DBLP:conf/fmcad/GoliaSCM21}, QuickSampler~\cite{DBLP:conf/icse/DutraLBS18}, UniGen~\cite{DBLP:conf/cav/SoosGM20} and KUS~\cite{DBLP:conf/lpar/SharmaGRM18}. Notice these samplers cannot sample SAT solutions from a non-uniform distribution. We include them in the learning experiments as a comparison, and exclude them in the weighted sampling experiment (in Fig.~\ref{fig:weighted_sample}).
\end{itemize}

\subsubsection{Metrics} In terms of evaluation metrics, we consider:
\begin{itemize}
    \item Training time per iteration, which computes  the average time for every  learning method  to finish one iteration.
    \item Validness, that is the percentage of generated solutions that satisfy the given constraints $\mathcal{C}$.
    \item  Mean Averaged Precision (MAP$@10$), which is the percentage that the solutions in the training set $\mathcal{D}$ reside among the top-10 \textit{w.r.t.} likelihood score. The higher the MAP@10 scores, the better the model generates structures closely resembling those in the training set.
    \item $\log$-likelihood of the solutions in the training set $\mathcal{D}$ (in Eq.~\ref{eq:log-likelihood}). The model that attains the highest $\log$-likelihood learns the closest  distribution to the training set. 
    \item Approximation error of  $\nabla \log Z_{\mathcal{C}}(\theta)$, which is the  $L_1$ distance between the exact value $ \nabla \log Z_{\mathcal{C}}(\theta)$ and the approximated value given by the sampler.
\end{itemize}
See Appendix D for detailed settings of baselines and evaluation metrics, as well as the following task definition, dataset construction, and potential function definition.

\subsection{Random $K$-SAT Solutions with Preference}
\subsubsection{Task Definition \& Dataset}  This task is to learn to generate solutions to a $K$-SAT problem. 
We are given a training set $\mathcal{D}$ containing solutions to a corresponding CNF formula $c_1\wedge\ldots\wedge  c_L$.
Note that not all solutions are equally likely to be presented in $\mathcal{D}$. 
The \textit{learning} task is to maximize the log-likelihood of the assignments seen in the training set $\mathcal{D}$. Once learning is completed, the \textit{inference} task is to generate valid solutions that closely resemble those in $\mathcal{D}$ ~\cite{DBLP:conf/aiia/DodaroP19}. 
To generate the training set $\mathcal{D}$, we use CNFGen~\cite{DBLP:conf/sat/LauriaENV17} to generate the random $K$-SAT problem and use Glucose4 solver to generate random valid solutions~\cite{DBLP:conf/sat/IgnatievMM18}.

\begin{table*}[!t]
    \centering
    \begin{tabular}{r|rrrrrrrrr}
    \hline
      Problem &\multicolumn{9}{c}{{(a) Training time per iteration} (Mins) ($\downarrow$)} \\ \cline{2-10}
     size & \nls& XOR & WAPS & WeightGen & CMSGen & KUS &QuickSampler & Unigen & Gibbs\\ \hline
$10$   & \textbf{0.13}  & $26.30$   & $1.75$ & $0.64$     & $0.22$  & 0.72& 0.40 & 0.66  & 0.86 \\
$20$   & \textbf{0.15}  & $134.50$  & $3.04$ & {T.O.}    & 0.26  & 0.90& 0.30 & 2.12  & 1.72 \\
$30$   & \textbf{0.19}  & $1102.95$ & $6.62$ & {T.O.}     & 0.28  & 2.24& 0.32 & 4.72  & 2.77 \\
$40$   & \textbf{0.23}  & T.O.   & 33.70 & {T.O.}    & 0.31  & 19.77 & 0.39 & 9.38  & 3.93 \\
$50$   & \textbf{0.24}  & T.O.   & 909.18& {T.O.}    & 0.33  & 1532.22  & 0.37 & 13.29 & 5.27 \\
$500$  & \textbf{5.99}  & T.O.   & {T.O.} & {T.O.}      & 34.17   & {T.O.} & {T.O.}  & {T.O.}   & $221.83$ \\
$1000$ & \textbf{34.01} & T.O.   & {T.O.} & {T.O.}      & $177.39$    & {T.O.} & {T.O.}  & {T.O.}   & $854.59$\\
\hline
&\multicolumn{9}{c}{(b) Validness of generated solutions ($\%$) ($\uparrow$)} \\ \hline
$10-50$ &  $\mathbf{100}$ & $\mathbf{100}$  &  $\mathbf{100}$  & $\mathbf{100}$  &  $\mathbf{100}$  & $\mathbf{100}$  &  $82.65$ & $\mathbf{100}$  & $90.58$\\
$500$ &   $\mathbf{100}$  & T.O. &  T.O. & T.O. &  $\mathbf{100}$  & T.O. &  $7.42$ & $\mathbf{100}$  & $54.27$\\
$1000$ &   $\mathbf{100}$  & T.O. &  T.O. & T.O. &  $\mathbf{100}$  & T.O. &  $0.00$ & $\mathbf{100}$  & $33.91$\\
\hline
 & \multicolumn{9}{c}{(c) Approximation error of  $\nabla\log Z_{\mathcal{C}}(\theta)$ ($\downarrow$)} \\ \hline
    10 & \textbf{0.10} & 0.21 & 0.12   & 3.58 & 3.96  & 4.08  & 3.93  & 4.16  & 0.69 \\
12 & \textbf{0.14} & 0.19 & 0.16   & 5.58 & 5.50  & 5.49  & 5.55  & 5.48  & 0.75 \\
14 & \textbf{0.15} & 0.25 & 0.19   & T.O. & 6.55  & 6.24  & 7.79  & 6.34  & 1.30 \\
16 & 0.16   & 0.25 & \textbf{0.15} & T.O. & 9.08  & 9.05  & 9.35  & 9.03  & 1.67 \\
18 & \textbf{0.18} & 0.30 & 0.23   & T.O. & 10.44 & 10.30 & 11.73 & 10.20 & 1.90 \\
\hline
    \end{tabular}
 \caption{Sampling efficiency and accuracy for learning $K$-SAT solutions with preferences. The proposed \nls is the most efficient (see ``Training Time Per Epoch'') and always generates valid assignments (see ``Validness'') with a small approximation error (see ``Approximation Error of Gradient'') against all baselines. T.O. means time out.}\label{tab:sampler}
\end{table*}

\subsubsection{Sampler's Efficiency and Accuracy}  Table~\ref{tab:sampler} shows the proposed \nls is an efficient sampler  that generates valid assignments, in terms of the training time for learning constrained MRF, approximation error for the gradient and  validness of the generated  assignments. In Table~\ref{tab:sampler}(a), \nls takes much less time for sampling against all the samplers and can train the model with the dataset of problem size $1000$ within an hour. In Table~\ref{tab:sampler}(b), \nls always generates valid samples. {The performance of} QuickSampler and Gibbs methods decreases when the problem size becomes larger.  In Table~\ref{tab:sampler}(c), \nls, XOR and WAPS are the three algorithms that can effectively estimate the gradient while the other algorithms incur huge estimation errors. Also, the rest methods are much slower than \nls. 

\noindent\textbf{Learning Quality}  Table~\ref{tab:learn-sat} demonstrates \nls-CD learns a more accurate  constrained MRF model by measuring the log-likelihood and MAP@10 scores. Note that baselines including Quicksampler, Weightgen, KUS, XOR and WAPS  timed out for the problem sizes we considered. Compared with the remaining baselines,  \nls attains the best log-likelihood and MAP@10 metric. 

\begin{table}[!t]
    \centering
    \scalebox{0.85}{
    \begin{tabular}{r|rrr|c}
    \hline
   & \multicolumn{4}{c}{{(a) $\log$-likelihood ($\uparrow$)}} \\
   \hline
     Problem & \multirow{3}{*}{\nls}  &\multirow{3}{*}{Gibbs} & \multirow{3}{*}{CMSGen} & Quicksampler\\
     size & & & &WeightGen,KUS\\ 
    &  &  &  & XOR, WAPS\\\hline
$100$ & $-49.16$ & $\mathbf{-36.36}$  & $-60.12$ &  \multirow{5}{*}{T.O.}\\
$300$ & $\mathbf{-52.61}$ & $-53.11$  & $-128.39$ &    \\
$500$ & $\mathbf{-196.47}$ & $-197.21$  & $-272.49$ &    \\
$700$ & $\mathbf{-238.60}$ & $-238.75$  & $-389.44$ &    \\
$1000$ & $\mathbf{-294.22}$ & $-296.33$  & $-532.85$ &   \\
\hline
 &\multicolumn{4}{c}{(b) MAP@10 (\%) ($\uparrow$) } \\ \hline
    $100$ &    $82.13$ &  $83.32$  & $\mathbf{86.34}$ &\multirow{5}{*}{T.O.} \\ 
    $300$ &  $\mathbf{66.37}$ & $64.42$ & $64.50$ & \\
    $500$ & $\mathbf{90.03}$ & $73.14$ & $70.67$ & \\
    $700$ & $\mathbf{69.74}$ & $\mathbf{69.74}$& $48.10$ &\\
    $1000$ &   $\mathbf{91.70}$   & $77.56$ & $78.72$ & \\
\hline
    \end{tabular}}
 \caption{The quality of learning outcomes for learning random $K$-SAT solutions with  preferences. \nls achieves the best likelihood and MAP@10 scores. T.O. is time out.}\label{tab:learn-sat}
\end{table}

\begin{figure}[!t]
    \centering
    \includegraphics[width=0.49\linewidth]{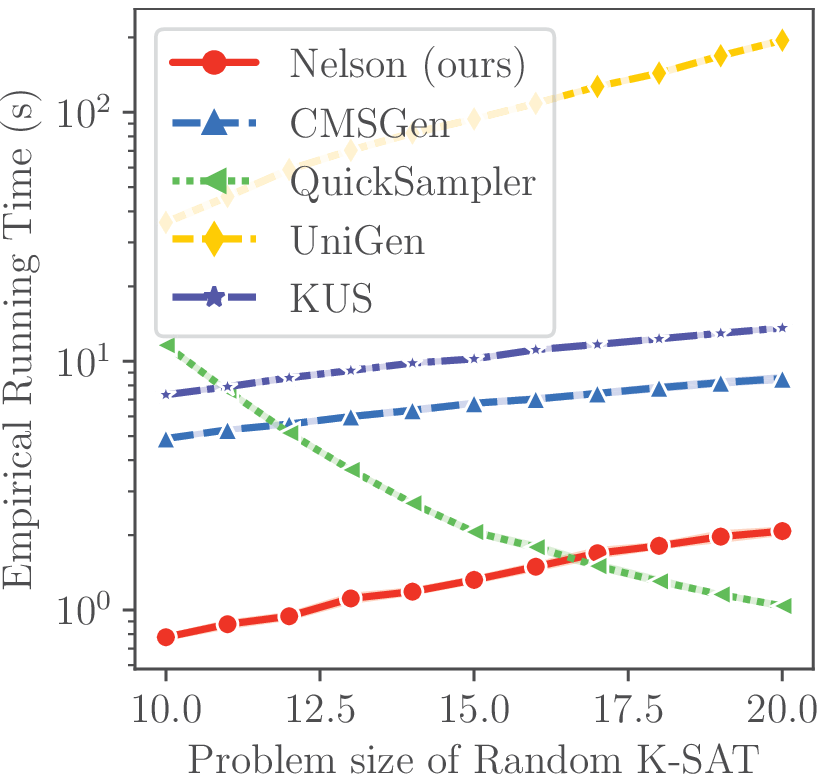}
    \includegraphics[width=0.50\linewidth]{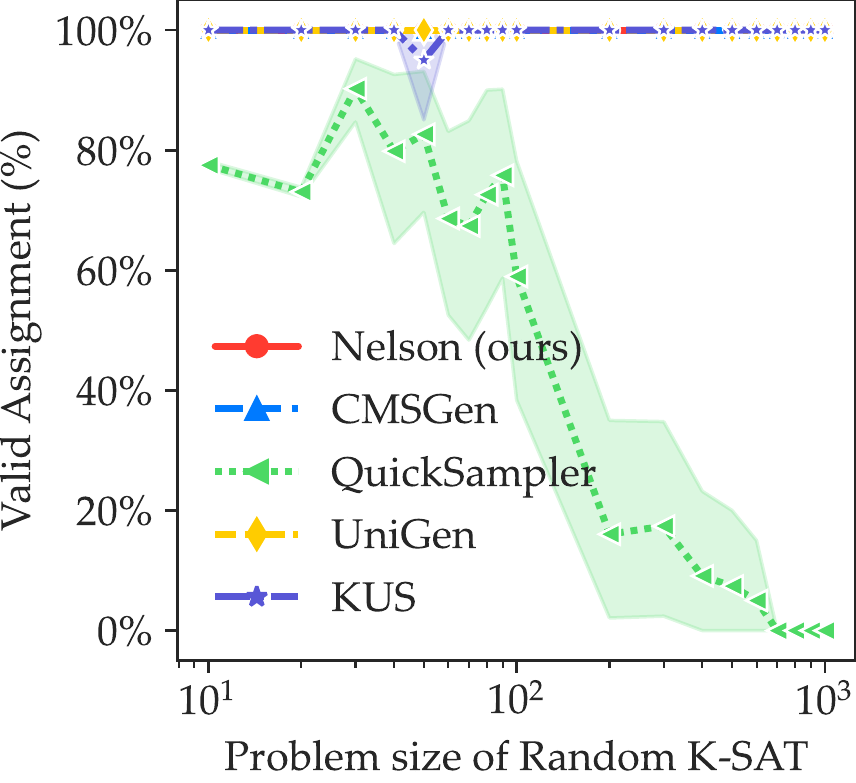}
    \caption{Running time and the percentage of valid structures sampled  uniformly at random from solutions of K-SAT problems. {Among all the problem sizes}, \nls always generate valid solutions  and is the most efficient sampler.}
    \label{fig:uniform_sample}
\end{figure}

\begin{figure}[!t]
\begin{minipage}{.53\linewidth}
  \centering
  \includegraphics[width=1\linewidth]{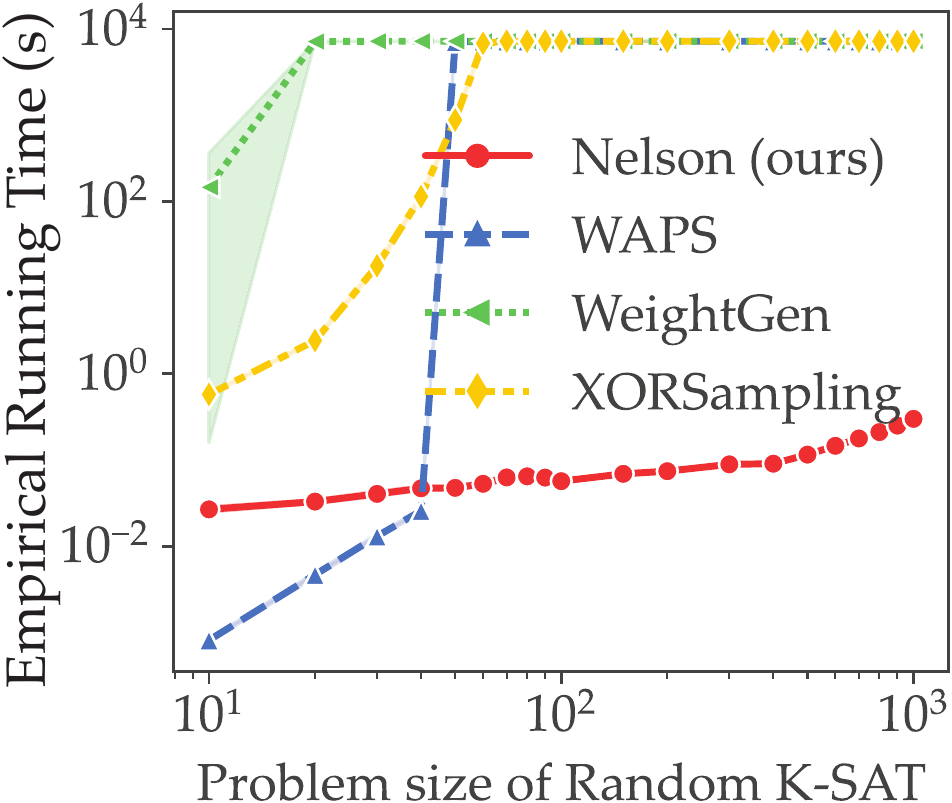}
\end{minipage}%
\begin{minipage}{.47\linewidth}
\raggedleft
   \includegraphics[width=0.99\linewidth]{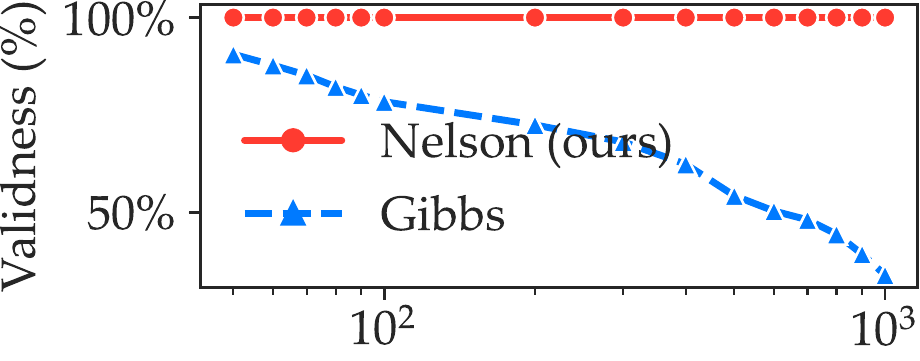}\\
  \includegraphics[width=0.89\linewidth]{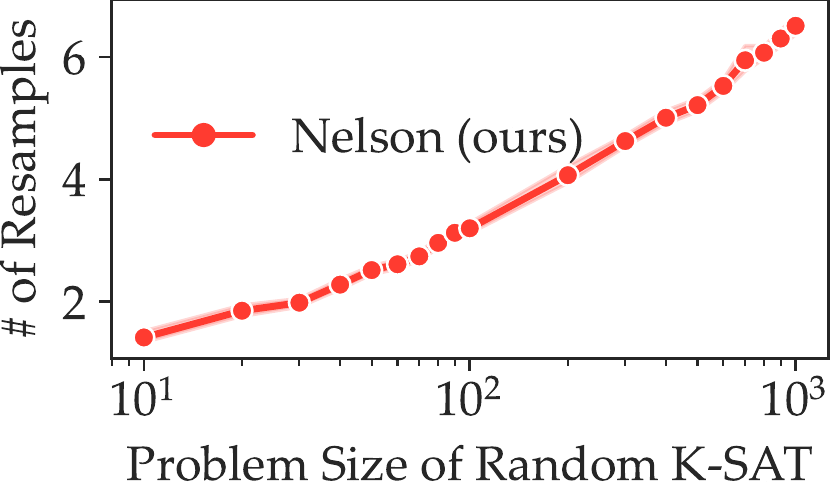}
\end{minipage}
    \caption{Running time, the percentage of valid solutions generated, and rounds of resampling for weighted sample generation of K-SAT solutions. {Among all the problem sizes}, \nls scales the best among all approaches and always generates valid solutions. }
    \label{fig:weighted_sample}
\end{figure}

\subsubsection{Abalation Study} We also evaluated the samplers' efficiency in isolation (not embedded in  learning). 
The sampling cases we considered are uniform and weighted (mainly following the experiment setting in~\citet{DBLP:conf/aaai/ChakrabortyM19}). 
In weighted sampling, the weights are specified by fixed values to the single factors in Eq.~\eqref{eq:constr_mrf_single}.
In the uniform sampling case in Fig.~\ref{fig:uniform_sample}, \nls and Quicksampler require much less time to draw  samples compared to other approaches. However, the solutions generated by  Quicksampler rarely satisfy constraints. 
In the weighted sampling case in Fig.~\ref{fig:weighted_sample}, \nls scales better  than all the competing samplers as the sizes of the $K$-SAT problems increase.

\begin{table}[!t]
    \centering
    \begin{tabular}{r|rrr}
    \hline
   Problem & \multicolumn{3}{c}{{(a) Training Time Per Epoch} (Mins) ($\downarrow$)} \\ \cline{2-4}
         size & \nls  &Gibbs & CMSGen  \\ 
        \hline
        $10$ &  $\mathbf{0.53}$ & $9.85$ & $0.69$ \\ 
        $20$ &  $\mathbf{0.53}$ & $80.12$ & $1.93$ \\ 
        $30$ &  $\mathbf{0.72}$ & $256.38$ & $3.65$ \\ 
        $40$ &  $\mathbf{0.93}$ & $777.01$ & $5.99$ \\ 
        $50$ &  $\mathbf{1.17}$ & T.O. & $9.08$ \\ 
        \hline
    &  \multicolumn{3}{c}{{(b) Validness of Orientations} ($\%$) ($\uparrow$)} \\ \hline
 $7$ & $\mathbf{100}$  &  $50.16$ & $\mathbf{100}$ \\   
$8$ & $\mathbf{100}$  &  $64.63$ & $\mathbf{100}$ \\ 
   $9$ & $\mathbf{100}$ & $47.20$ & $\mathbf{100}$  \\  
   $10$ & $\mathbf{100}$ & $62.60$ & $\mathbf{100}$  \\ 
   $11$ & $\mathbf{100}$     & $84.95$ & $\mathbf{100}$\\ 
    \hline
    &  \multicolumn{3}{c}{(c) Approximation Error of  $\nabla\log Z_{\mathcal{C}}(\theta)$  ($\downarrow$)} \\ \hline
    $5$ & $\mathbf{0.01}$ & $0.09$  & $0.21$ \\
$7$ & $\mathbf{0.05}$ &$0.08$ &  $2.37$ \\
$9$   & $\mathbf{0.03}$ & $0.11$  & $2.37$  \\
$11$  & $\mathbf{0.04}$ &	 $0.17$  & $8.62$ \\
$13$ & $\mathbf{0.05}$ &$0.28$	&  $11.27$ \\
\hline
&  \multicolumn{3}{c}{(d) MAP@10 (\%) ($\uparrow$) } \\ \hline
$10$ & $61.14$ & $60.01$ & $\mathbf{64.56}$\\
$20$ & $\mathbf{55.26}$ & $55.20$ & $47.79$\\
$30$ & $\mathbf{100.00}$ & $96.29$  &$\mathbf{100.00}$\\
$40$ & $\mathbf{40.01}$  & $39.88$ & $38.90$\\
$50$ & $\mathbf{46.12}$ & T.O. & $42.11$ \\
\hline
    \end{tabular}
\caption{Sample efficiency and learning performance of the sink-free orientation task. \nls is the most efficient (see Training Time Per Epoch) and always generates valid assignments (see Validness), has the smallest error approximating gradients, and has the best learning performance (see MAP@10) among all baselines.}\label{tab:sink-free}
\end{table}

\subsection{Sink-Free Orientation in Undirected Graphs}
\noindent\textbf{Task Definition \& Dataset}  A \textit{sink-free} orientation of an undirected graph is a choice of orientation for each arc such that every vertex has at least one outgoing arc~\cite{DBLP:journals/combinatorics/CohnPP02}. This task has wide applications in robotics routing and IoT network configuration~\cite{takahashi2009communication}. Even though finding a sink-free orientation is tractable, sampling a sink-free orientation from the space of all orientations is still \#P-hard. Given a training set of preferred orientations $\mathcal{D}$ for the graph, the \textit{learning} task is to maximize the log-likelihood of the orientations seen in the training set. The \textit{inference} task is to generate valid orientations that resemble those in the training set. To generate the training set, we use the Erd\H{o}s-R\'enyi random graph  from the NetworkX library. The problem size is characterized by the number of vertices in the graph. The baselines we consider are CD-based learning with Gibbs sampling and CMSGen.

\subsubsection{Learning Quality} In Table~\ref{tab:sink-free}(a), we show the proposed \nls method takes much less time to train MRF for one epoch  than the competing approaches. Furthermore, in Table~\ref{tab:sink-free}(b), \nls and CMSGen generate $100\%$ valid orientations of the graph while the Gibbs-based model does not.  
Note the constraints for this task satisfy Condition~\ref{cond:extreme}, hence \nls sampler's performance is guaranteed by Theorem~\ref{th:product-dist}.
In Table~\ref{tab:sink-free}(c), \nls attains the smallest approximation error for the gradient (in Eq.~\ref{eq:gradient}) compared to baselines. Finally, \nls  learns a higher MAP@10 than CMSGen. The Gibbs-based approach times out for problem sizes larger than $40$. In summary, our \nls is the best-performing algorithm for this task.

\subsection{Learn Vehicle Delivery Routes}
\subsubsection{Task Definition \&  Dataset} Given a set of locations to visit, the task is to generate a sequence to visit these locations in which each location is visited once and only once and the sequence closely resembles the trend presented in the training data.  The training data are such routes collected in the past.  
The dataset is constructed from TSPLIB, which consists of $29$ cities in Bavaria, Germany. The constraints for this problem do not satisfy Condition~\ref{cond:extreme}.  We still apply the proposed method to evaluate if the \nls algorithm can handle those general hard constraints.

\begin{figure}[!t]
    \centering
    \includegraphics[width=0.495\linewidth]{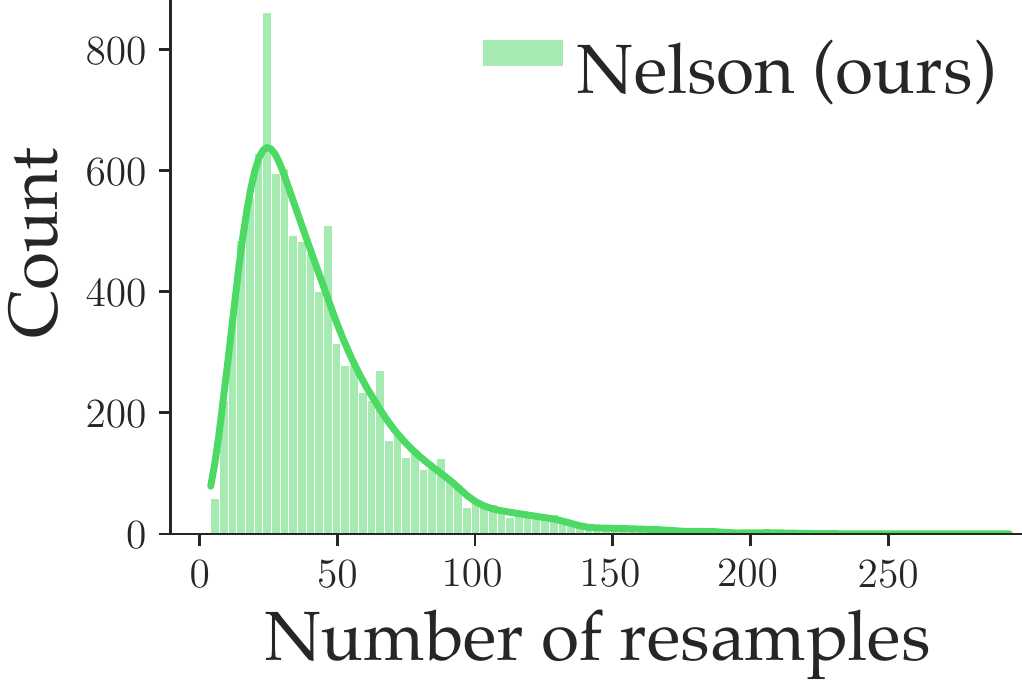}
    \includegraphics[width=0.495\linewidth]{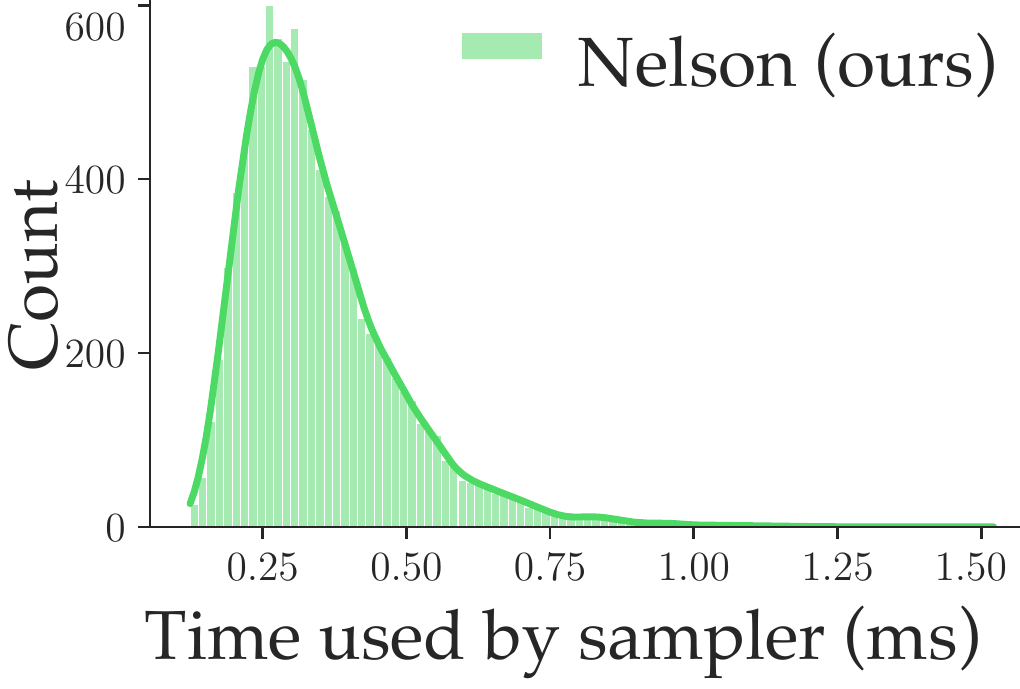}
    \caption{Frequency histograms for the number of resample and the total time of \nls method for uniformly sampling visiting paths for vehicle routing problem.}
    \label{fig:route}
\end{figure}

In Fig.~\ref{fig:route}, we see  \nls can obtain samples of this delivery problem efficiently. We measure the number of resamples taken as well as the corresponding time used  by the \nls method. \nls takes roughly 50 times of resamples with an average time of $0.3$ seconds to draw a batch (the batch size is $100$) of valid visiting sequences.

\section{Conclusion}
In this research, we present \nls, which  embeds a sampler based on \lovasz Local Lemma into the contrastive divergence learning of Markov random fields. The embedding is fully differentiable. This approach allows us to learn generative models over constrained domains, which presents significant challenges to other state-of-the-art models. 
We also give sound proofs of the performance of the LLL-based sampler.
Experimental results on several real-world domains reveal that \nls learns to generate 100\% valid structures, while baselines either time out or cannot generate valid structures. 
\nls also outperforms other approaches in the running times and in various learning metrics.

\section{Acknowledgments}
We thank all the reviewers for their constructive comments. This research was supported by
NSF grants IIS-1850243, CCF-1918327. 

\bibliography{reference,fan}
\appendix

\newpage
\onecolumn

\section{Probability Distribution  of Algorithm~\ref{alg:lll-sampler}}\label{appendix:prob-dist}
\subsection{Definitions and Notations}
This section is for the proofs related to the probability distribution in the proposed Algorithm~\ref{alg:lll-sampler}. For convenience, commonly used notations are listed in Table~\ref{tab:quantifier}. We make some slight changes to some notations that appear in the main paper to make sure they are consistent and well-defined in this proof.

Similar to the previous analysis~\cite{DBLP:journals/jacm/GuoJL19,DBLP:journals/arxiv/jerrum2021}, we begin by introducing the concept ``dependency graph'' (in Definition~\ref{def:dep-graph}) for the constraints $\mathcal{C}$.

\begin{definition}[Dependency Graph] \label{def:dep-graph}
The dependency graph $G=(\mathcal{C},E)$, where the vertex set is the set of constraints $\mathcal{C}$.  Two vertices $c_i$ and $c_j$ are connected with an edge $(c_i,c_j)\in E$ if and only if they are defined on at least one common random variable, i.e., $\var(c_i)\cap\var(c_j) \neq \emptyset$. 
\end{definition}

For keeping track of the whole sampling procedure, we need the concept ``sampling record''(in Definition~\ref{def:record})~\cite{DBLP:journals/jacm/GuoJL19}, which are the broken constraints at every round in Algorithm~\ref{alg:lll-sampler}. It is also known as the witness tree in~\citet{DBLP:journals/jacm/MoserT10}.  This allows us to check the constraint satisfaction of the assignment at every round.

Under Condition~\ref{cond:extreme}, for any edge in the dependency graph $(c_i,c_j) \in E$,   either $\mathbf{1}(x,c_i)=0$ or $\mathbf{1}(x,c_j)=0$ for all $x\in \mathcal{X}$.
In other words, two constraints with shared related variables, representing two \textit{adjacent vertices} in the dependency graph $G$, are not broken simultaneously. Thus, the constraints in record $S_t$ form an \textit{independent set}\footnote{A set of vertices with no two adjacent vertices in the graph.} over the dependency graph under Condition~\ref{cond:extreme}.

\begin{definition}[Sampling Record] \label{def:record} 
Given dependency graph $G(\mathcal{C},E)$, let $X_t=x$ be  one possible assignment obtained at round $t$ of Algorithm~\ref{alg:lll-sampler}. Let $S_t\subseteq \mathcal{C}$ be the set of vertices in graph $G$ (subset of constraints) that $x$ violates
\begin{align}
    S_t = \{c_i|c_i\in \mathcal{C}\text{ and } \mathbf{1}(x,c_i)=0\},
\end{align}
where indicator function $\mathbf{1}(x,c_i)=0$ implies  $x$ violates constraint $c_i$  at round $t$.
Define the sampling record as the sequence of violated constraints  $S_1,\ldots,S_t$ throughout the execution.
\end{definition}

At round $t$  ($t \geq 1$) of Algorithm~\ref{alg:lll-sampler}, suppose the violated constraints is $S_t\subseteq \mathcal{C}$. The constraints that are not adjacent to $S_t$ in the dependency graph are still satisfied after re-sample. The only possible constraints that might be broken after the re-sample operation are among $S_t$ itself, or those constraints directly connected to  $S_t$ in the dependency graph. Therefore,
\begin{equation*}
S_{t+1} \subset \Gamma(S_t),    \qquad \text{ for all } t\ge 1.
\end{equation*}
where $\Gamma(S_t)$ is the set of vertices of $S_t$ and its adjacent neighbors
in the dependency graph $G$ (see Table~\ref{tab:quantifier}).  When Algorithm~\ref{alg:lll-sampler} terminates at round $T+1$, no  constraints are violated anymore, i.e., $s_{T+1}=\emptyset$.  To summarize the above discussion on a sampling record by Algorithm~\ref{alg:lll-sampler}, we have the following Claim~\ref{claim:record}.
\begin{claim}~\label{claim:record}
Under Condition~\ref{cond:extreme}, a potential sampling record of length $T+1$ by the Algorithm~\ref{alg:lll-sampler} is a sequence of independent sets: $S_1, S_2, \ldots,S_T,\emptyset$ with 
\begin{enumerate}
    \item $S_{t+1} \subseteq \Gamma(S_{t})$ and $S_t\neq\emptyset$, for $1 \leq t\leq T$; 
    \item $s_{T+1}=\emptyset$.
\end{enumerate}
\end{claim}

\begin{table}[!t]
    \centering
    \caption{Summary of all the notations used in the theoretical analysis of Algorithm~\ref{alg:lll-sampler}.}\label{tab:quantifier}
    \begin{tabular}{ll}
    \hline
    Notation& Definition \\
    \hline
        $X=\{X_i\}_{i=1}^n$ & set of discrete random variables \\
        \hline
        $x\in\mathcal{X}$ & possible assignments for variables $X$ \\
        \hline
        $x_i\in\mathcal{X}_i$ & variable $X_i$ can take all values in $\mathcal{X}_i$\\
        \hline
        $\mathcal{C}=\{c_j\}_{j=1}^{m}$ & given constraints \\
        \hline
        $S_t\subseteq\mathcal{C}$& subset of constraints violated at round  $t$ of Algorithm~\ref{alg:lll-sampler} \\
        \hline
        $G(\mathcal{C},E)$& the dependency graph (in Definition~\ref{def:dep-graph})\\
        \hline
        $\var(c_j)$& the indices of domain variables that are related to constraint $c_i$\\
        \hline
        $\var(S_t)$& the indices for domain variables that are related to constraints $S_t$\\
        \hline
        $\Gamma(c_j)$ &$c_j$ and its direct neighbors in the dependency graph\\
        \hline
        $\Gamma(S_t)$ &$S_t$ and direct neighbors of $S_t$ in the      dependency graph\\
        \hline
        $\mathcal{C}\backslash \Gamma(S_t)$ &all constraints in $\mathcal{C}$ but not in $\Gamma(S_t)$ \\
        \hline
        $S_1,\ldots,S_{T},\emptyset$ & a sampling record of Algorithm~\ref{alg:lll-sampler}  (in Definition~\ref{def:record}) \\
        \hline
        $\mathbf{1}(x,c_i)$ & indicator function that evaluates if    assignment $x$ satisfies constraint $c_i$ \\
        \hline
        $\mathbf{1}(x,S_t)$ & indicator function that evaluates if assignment $x$ satisfies constraints in $S_t$ \\
        \hline
        $\mathbf{1}(x,\mathcal{C})$ & indicator function that evaluates if  assignment $x$ satisfies all constraints $\mathcal{C}$ \\
        \hline
        $P_{\theta}(X|\mathcal{C}\backslash\Gamma(S_t))$& see Definition~\ref{def:general-cmrf}\\
        \hline
        $\p(X|S_t)$ & see Definition~\ref{def:prs-prob}\\
    \hline
    \end{tabular}
\end{table}

\subsubsection{Extra Notations related to Constrained MRF}
The constrained MRF model over constraints set $\mathcal{C}$ is defined as:
\begin{align*}
P_{\theta}(X=x|\mathcal{C})&=\frac{\exp(\sum_{i=1}^n\theta_i x_i)\mathbf{1}(x,\mathcal{C})}{\sum_{x'\in\mathcal{X}}\exp(\sum_{i=1}^n\theta_i x'_i)\mathbf{1}(x',\mathcal{C})}
\end{align*}
where the partition function only sums over valid assignments in $\mathcal{X}$. Note that $C(x)$ in Equation~\eqref{eq:constr_mrf_single} is the same as $\mathbf{1}(x',\mathcal{C})$ in the above equation. We slightly change the notations for consistency in this proof. Also notice that the output distribution can no longer be factorized after constraints are enforced, since the partition function cannot be factorized. Our task is to draw samples from this distribution. 

To analyze the intermediate steps in Algorithm~\ref{alg:lll-sampler}, we further need to define the following notations.
\begin{definition}  \label{def:general-cmrf}
The constrained MRF distribution for constraints $\mathcal{C}\backslash \Gamma(S_t)$ is
\begin{equation*}\footnotesize
\begin{aligned}
P_{\theta}(X=x|\mathcal{C}\backslash \Gamma(S_t))&=\frac{\exp(\sum_{i=1}^n\theta_i x_i)\mathbf{1}(x,\mathcal{C}\backslash \Gamma(S_t))}{\sum_{x'\in\mathcal{X}}\exp(\sum_{i=1}^n\theta_i x'_i)\mathbf{1}(x',\mathcal{C}\backslash \Gamma(S_t))}
\end{aligned}
\end{equation*}
\end{definition}

\begin{definition}  \label{def:prs-prob}
At round $t$ of Algorithm~\ref{alg:lll-sampler}, assume $S_t\subseteq\mathcal{C}$ are the set of broken constraints,  Define $\p(X_{t+1}=x|S_1,\ldots, S_t)$ to be the probability of obtaining a new assignment $x$ after we re-sample random variables indexed by $\var(S_t)$.
\end{definition}

\subsection{Ratio Property Lemma}

\begin{lemma}[Ratio Property] \label{lem:ratio-prob}
Under Condition~\ref{cond:extreme}, assume Algorithm~\ref{alg:lll-sampler} is at round $t$. Conditioning on observing one possible sampling record $S_1,\ldots,S_t$, Algorithm~\ref{alg:lll-sampler} step 4 will re-sample variables in $\var(S_t)$ at round $t+1$. Let $x,x'\in\mathcal{X}$ be two possible assignments after this re-sample.  The probability ratio  of obtaining these two results equals that under constrained MRF
$P_{\theta}(x|\mathcal{C}\backslash \Gamma(S_t))$:
\begin{equation} \label{eq:ratio}
\frac{\p(X_{t+1}= x |S_1,\ldots, S_t)}{\p(X_{t+1}= x'|S_1,\ldots, S_t)} = \frac{P_{\theta}(X = x |\mathcal{C}\backslash \Gamma(S_t))}{P_{\theta}(X = x'|\mathcal{C}\backslash \Gamma(S_t))},
\end{equation} 
where $\p(X_{t+1}= x |S_1,\ldots, S_t)$ is the probability of Algorithm~\ref{alg:lll-sampler} step 4 produces assignment $x$ at round $t+1$, conditioning on the observed record $S_1,\ldots, S_t$ and re-sample $\var(S_t)$. $P_{\theta}(X = x |\mathcal{C}\backslash \Gamma(S_t))$ is the constrained MRF (for the constraints  $\mathcal{C}\backslash \Gamma(S_t)$) probability on assignment $x$.
\begin{proof}
During the intermediate step of the algorithm, assume the set of constraints $S_t$ are violated. We want to re-sample variables indexed by $\var(S_t)$, so  variables indexed by $\var(\mathcal{C}\backslash\Gamma(S_t))$ won't change assignments. Also, because $\Gamma(S_t)$ is the largest possible set of constraints that can be infected by the re-sample, constraints $\mathcal{C}\backslash \Gamma(S_t)$ are still satisfied after the re-sample. 

At round $t$,  we re-sample variables in $\var(S_t)$ according to step 4 in Algorithm~\ref{alg:lll-sampler}, we thus have:
\begin{equation*}
\p(X^{t+1}_{\var(S_t)}=x_{\var(S_t)}|S_1 \ldots S_t)=\underset{{i\in\var(S_t)}}{\prod}\frac{\exp(\theta_i x_i)}{\underset{x'_i\in \mathcal{X}_i}{\sum}\exp(\theta_ix'_i)}.
\end{equation*}
Here the notation $X^{t+1}_{\var(S_t)}=x_{\var(S_t)}$ means $X_i=x_i$ for $i\in\var(S_t)$ at round $t$. For any two possible assignments $x,x'$ after the re-sample,
\begin{equation*}
x_i=x'_i,\quad \text{ for }i\in \{1,\ldots,n\}\backslash \var(S_t)
\end{equation*}
since the rest variable's assignments are kept the same after re-sample. Thus, we can have ratio:
\begin{equation}\label{eq:ratio-nelson}
\frac{\p(X_{t+1}=x|S_1,\ldots,S_t)}{\p(X_{t+1}=x'|S_1,\ldots, S_t)}=\frac{\exp(\sum_{i\in\var(S_t)}\theta_ix_i)}{\exp(\sum_{i\in\var(S_t)}\theta_ix'_i)} =\frac{\exp(\sum_{i\in\var(\Gamma(S_t))}\theta_ix_i)}{\exp(\sum_{i\in\var(\Gamma(S_t))}\theta_ix'_i)}.
\end{equation}
For the last step, since every assignment outside $\var(S_t)$ is not changed, we can enlarge the index set of summation to $\Gamma(S_t)$ by multiplying
\begin{align*}
1 = \frac{\exp(\sum_{i\in\var(\Gamma(S_t)) \backslash \var(S_t)}\theta_ix_i)}{\exp(\sum_{i\in\var(\Gamma(S_t))\backslash \var(S_t)}\theta_ix'_i)}.
\end{align*}
After re-sample, we knows that $x$ must satisfy the constraints $\mathcal{C}\backslash\Gamma(S_t)$. Thus, the probability of this $x$ conditioned on constraints $\mathcal{C}\backslash\Gamma(S_t)$ holding in the constrained MRF model is:
\begin{equation*}
P_{\theta}(X=x|\mathcal{C} \backslash \Gamma(S_t))
=\frac{\exp(\sum_{i=1}^n\theta_i x_i)\mathbf{1}(x,\mathcal{C}\backslash \Gamma(S_t))}{\sum_{x'\in\mathcal{X}} \exp(\sum_{i=1}^n\theta_i x'_i)\mathbf{1}\left(x',\mathcal{C}\backslash \Gamma(S_t)\right)}=\frac{\exp(\sum_{i=1}^n\theta_i x_i)}{\sum_{x'\in\mathcal{X}} \exp(\sum_{i=1}^n\theta_i x'_i)\mathbf{1}\left(x',\mathcal{C}\backslash \Gamma(S_t)\right)}.
\end{equation*}
In the constrained MRF model (for constraints  $\mathcal{C}\backslash\Gamma(S_t)$), the ratio of these two probabilistic assignments $x,x'$ is:
\begin{equation}\label{eq:ratio-cmrf}
\begin{aligned}
\frac{P_{\theta}(X=x|\mathcal{C}\backslash \Gamma(S_t))}{P_{\theta}(X=x'|\mathcal{C}\backslash\Gamma(S_t))}=\frac{\exp(\sum_{i\in \var(\Gamma(S_t))}\theta_ix_i)}{\exp(\sum_{i\in \var(\Gamma(S_t))}\theta_ix'_i)},
\end{aligned}
\end{equation}
because the $x_i$ outside $\var(\Gamma(S_t))$ remains the same. Note that $x,x'$ are two possible assignments produced according to to step 4 in Algorithm~\ref{alg:lll-sampler} at round $t$. 
Combining Equation~\eqref{eq:ratio-nelson} and Equation~\eqref{eq:ratio-cmrf}, we conclude that:
\begin{equation*}
\begin{aligned}
\frac{\p(X_{t+1}=x|S_1,\ldots,S_t)}{\p(X_{t+1}=x'|S_1,\ldots,S_t)}=\frac{P_{\theta}(X=x|\mathcal{C}\backslash \Gamma(S_t))}{P_{\theta}(X=x'|\mathcal{C}\backslash\Gamma(S_t))}.
\end{aligned}
\end{equation*}
The proof is finished. 
\end{proof}
\end{lemma}

\subsection{Proof of Theorem~\ref{th:product-dist}}
Suppose the re-sampling process terminates at round $T+1$ and we obtain a valid sample $x$. Upon the termination of Algorithm~\ref{alg:lll-sampler}, all the constraints are satisfied. So we have: $S_{T+1}=\emptyset$. In other words, $\mathbf{1}(x,\mathcal{C})=1$. 

Let $x,x'$ be two possible valid assignments produced at round $T+1$ by the Algorithm~\ref{alg:lll-sampler}. Using the analysis in Lemma~\ref{lem:ratio-prob}, we can still have:
\begin{equation*}
\frac{\p(X_{T+1}=x|S_1,\ldots, S_T)}{\p(X_{T+1}=x'|S_1,\ldots, S_T)}=\frac{\exp(\sum_{i\in\var(S_T)}\theta_ix_i)}{\exp(\sum_{i\in\var(S_T)}\theta_ix'_i)}.
\end{equation*}
The probability of this $x$ in the constrained MRF model  (for constraints  $\mathcal{C}$) is:
\begin{equation*}
P_{\theta}(X=x|\mathcal{C})=\frac{\exp(\sum_{i=1}^n\theta_i x_i)\mathbf{1}(x,\mathcal{C})}{\sum_{x'\in\mathcal{X}}  \exp(\sum_{i=1}^n\theta_i x'_i)\mathbf{1}\left(x',\mathcal{C}\right)}
=\frac{\exp(\sum_{i=1}^n\theta_i x_i)}{\sum_{x'\in\mathcal{X}}  \exp(\sum_{i=1}^n\theta_i x'_i)\mathbf{1}\left(x',\mathcal{C}\right)}.
\end{equation*}
Then we conclude that:
\begin{equation*}
\frac{\p(X_{T+1}=x|S_1,\ldots, S_T)}{\p(X_{T+1}=x'|S_1,\ldots, S_T)}=\frac{P_{\theta}(X=x|\mathcal{C})}{P_{\theta}(X=x'|\mathcal{C})}.
\end{equation*}
Note that this ratio property holds for all the possible sampling records $S_1,\ldots, S_T,\emptyset$.

\subsubsection{Summation of All Possible Sampling Records} Define $\p(S_1,\ldots,S_T)$ to be the probability of observing record $S_1,\ldots,S_T$ by Algorithm~\ref{alg:lll-sampler}.
For any possible sampling record $S_1,\ldots,S_{T}, \emptyset$, the ratio property still holds:
\begin{equation*}
\frac{\p(X_{T+1}=x|S_1,\ldots S_T)\p(S_1,\ldots,S_T)}{\p(X_{T+1}=x'|S_1,\ldots,S_T)\p(S_1,\ldots,S_T)}=\frac{P_{\theta}(X=x|\mathcal{C})}{P_{\theta}(X=x'|\mathcal{C})}
\end{equation*}
where the term $\p(S_1,\ldots,S_T)$ on the Left-hand-side (LHS) is actually the same.
After we summarize over all possible sampling records $S_1,\ldots,S_{T},\emptyset$, the ratio property still holds. Let $\p(X_{T+1}=x)$ be the probability of obtaining one valid assignment $x$ by the execution of Algorithm~\ref{alg:lll-sampler}.
\begin{equation} \label{eq:ratio-final}
\frac{\p(X_{T+1}=x)}{\p(X_{T+1}=x')}=\frac{\sum_{S_1,\ldots,S_T}\p(X_{T+1}=x|S_1,\ldots,S_T)\p(S_1,\ldots,S_T)}{\sum_{S_1,\ldots,S_T}\p(X_{T+1}=x'|S_1,\ldots,S_T)\p(S_1,\ldots,S_T)}=\frac{P_{\theta}(X=x|\mathcal{C})}{P_{\theta}(X=x'|\mathcal{C})}
\end{equation}

\subsubsection{Sample Space Analysis At Termination}
We need one more statement to show Theorem~\ref{th:product-dist} holds. Let $\mathcal{X}_{\text{LLL}}$ be the set of all possible  assignments $x$ that can be generated by Algorithm~\ref{alg:lll-sampler}:
\begin{equation*}
\mathcal{X}_{\text{LLL}}=\bigcup_{S_1\ldots S_T}\{x|\p(X_{T+1}=x|S_1\ldots S_T) \neq 0 \text{ and } \p(S_1\ldots S_T)\neq 0\}.
\end{equation*}
{where $ \p(S_1\ldots S_T)\neq 0$ means $S_1,\ldots,S_T$ is a possible record. $\p(X_{T+1}=x|S_1\ldots S_T) \neq 0$ means it is possible to obtain $x$ given the record $S_1,\ldots,S_T$.}

Let $\mathcal{X}_{\mathcal{C}}$ be the set of assignments $x$ that satisfy all the constraints in the constrained MRF (for constraints $\mathcal{C}$):
\begin{equation*}
\mathcal{X}_{\mathcal{C}}=\{x|P_{\theta}(X=x|\mathcal{C}) \neq 0, \text{ for all }x\in\mathcal{X}\}.
\end{equation*}

\begin{lemma}\label{lem:space}
$\mathcal{X}_{\text{LLL}}\subseteq\mathcal{X}_{\mathcal{C}}$ and $\mathcal{X}_{\mathcal{C}}\subseteq\mathcal{X}_{\text{LLL}}$, thus $\mathcal{X}_{\text{LLL}}=\mathcal{X}_{\mathcal{C}}$.
\begin{proof}
When Algorithm~\ref{alg:lll-sampler} terminates, it only produces valid assignments; thus, we must have: $\mathcal{X}_{\text{LLL}}\subseteq\mathcal{X}_{\mathcal{C}}$. On the other hand,   there is always a non-zero probability that Algorithm~\ref{alg:lll-sampler} will generate every valid assignment $x\in\mathcal{X}_\mathcal{C}$, which implies that $\mathcal{X}_{\mathcal{C}}\subseteq\mathcal{X}_{\text{LLL}}$. Therefore we can conclude that $\mathcal{X}_{\text{LLL}}=\mathcal{X}_{\mathcal{C}}$.
\end{proof}
\end{lemma}
Lemma~\ref{lem:space} show that the two distributions have the same sample space when Algorithm~\ref{alg:lll-sampler} terminates. What's more, Equation~\eqref{eq:ratio-final} shows they have the same probability ratio for any possible valid assignments $x,x'$. This shows that the execution of the Algorithm~\ref{alg:lll-sampler} is a random draw from the constrained MRF distribution $P_{\theta}(X=x|\mathcal{C})$. The proof of Theorem~\ref{th:product-dist} is finished. 

\subsection{Difference to the Original Proof} 
 The main difference in the above proof to the existing proof in~\cite[Lemma 7]{DBLP:journals/jacm/GuoJL19} is that: We  show Lemma~\ref{lem:ratio-prob} that characterizes the proportional ratio of getting different assignments of variables, which is more general than the descriptive proof for \citet[Lemma 7]{DBLP:journals/jacm/GuoJL19}.

\subsection{{A Running Example in View of Markov Chain}}
{We dedicate this section to demonstrate the execution of Algorithm~\ref{alg:lll-sampler} with Example~\ref{example:matrix}. Algorithm~\ref{alg:lll-sampler} can be viewed as a Markov chain, so  we will show the probability of obtaining valid samples is unbiased by running thousands of steps of the Markov chain. }
The constraints are $\mathcal{C}=\{c_1=(X_1\vee X_2), c_2=(\neg X_1 \vee X_3)\}$. We use the assignment of all the variables as the states $s_1,\ldots, s_8$ in the rounds of Algorithm~\ref{alg:lll-sampler}.
\begin{equation}
\begin{aligned}
s_1&=(X_0=0,X_1=0,X_2=0) \\
s_2&=(X_0=0,X_1=0,X_2=1) \\
s_3&=(X_0=0,X_1=1,X_2=0) \\
s_4&=(X_0=0,X_1=1,X_2=1) \\
s_5&=(X_0=1,X_1=0,X_2=0) \\
s_6&=(X_0=1,X_1=0,X_2=1) \\
s_7&=(X_0=1,X_1=1,X_2=0) \\
s_8&=(X_0=1,X_1=1,X_2=1) \\
\end{aligned}
\end{equation}
Here $s_1,s_2,s_3,s_4$ correspond to \textit{valid} assignments of variables with respect to the constraints $\mathcal{C}$ and $s_5,s_6,s_7,s_8$ correspond to \textit{invalid} assignments of variables, that requires resampling.

For simplicity, we consider the uniform setting where $\theta_1=\theta_2=\theta_3$. The goal is to sample every valid assignment with equal probability. Therefore, the probability for every variable is: 
\begin{equation*}
P(X_i)=\begin{cases}
\frac{1}{2} &\text{ for variable } X_i\text{ taking value }1 \\
\frac{1}{2} &\text{ for variable } X_i\text{ taking value }0 \\
\end{cases}
\end{equation*}
for $i=1,2, 3$. Based on Algorithm~\ref{alg:lll-sampler}, we know the probability of transferring from $s_i$ to $s_j$ ($1\le i,j\le 8$). Thus we can construct the transition matrix between every state:
\begin{equation}
T=\begin{bNiceMatrix}[
  first-row,code-for-first-row=\normalsize,
  first-col,code-for-first-col=\normalsize,
]
 &s_1 &s_2 & s_3  & s_4 & s_5 & s_6 & s_7 & s_8 \\
s_1& \mathbf{1} & 0 & 0 & 0 & 0 & 0 & 0 & 0 \\
 s_2& 0 & \mathbf{1} & 0 & 0 & 0& 0 & 0 & 0\\
s_3& 0 & 0 &  \mathbf{1}  & 0 & 0& 0 & 0 & 0\\
s_4& 0 & 0 & 0 & \mathbf{1} & 0 & 0 & 0 & 0\\
s_5& \frac{1}{4} & \frac{1}{4} & \frac{1}{4} & 0& \frac{1}{4} & 0 & 0 & 0 \\
s_6 & 0 & \frac{1}{4} & 0 & 0& \frac{1}{4} & \frac{1}{4} & 0 & \frac{1}{4} \\
s_7 & \frac{1}{4} & 0 & \frac{1}{4} & \frac{1}{4}& 0 & 0 & \frac{1}{4}  & 0 \\
s_8& 0 & 0 & \frac{1}{4} & 0& 0 & \frac{1}{4}  & \frac{1}{4}  & \frac{1}{4}  \\
\end{bNiceMatrix}
\end{equation}
where $T_{ij}=T(s_i,s_j)$ denotes the transition probability from state $s_i$ to state $s_j$. 

{Taking state $s_5$ as an example, it violates constraint $C_2$ thus $X_2,X_3$ will be resampled. There are 4 possible assignments of  $X_2,X_3$, which corresponds to states $\{s_1,s_2,s_3,s_5\}$. Since each variable is resampled uniformly at random, the probability of transition from state $s_5$ to the states $\{s_1,s_2,s_3,s_5\}$ are $1/4$. The Algorithm~\ref{alg:lll-sampler} will terminate once it reaches states $\{s_1,s_2,s_3,s_4\}$, which corresponds to the (valid) state only transit to itself with probability 1. Thus we find $T(s_i,s_i)=1$ for $i=1,2,3,4$.}

For a randomly initialized assignment:
\begin{equation}
x=\begin{bNiceMatrix}[
  first-row,code-for-first-row=\normalsize,
  first-col,code-for-first-col=\normalsize,
]
 &s_1 &s_2 & s_3  & s_4 & s_5 & s_6 & s_7 & s_8 \\
 &\frac{1}{8} & \frac{1}{8} & \frac{1}{8} & \frac{1}{8} & \frac{1}{8} & \frac{1}{8} & \frac{1}{8} & \frac{1}{8} \\
\end{bNiceMatrix}
\end{equation}
that has an equal probability of being any state.
After executing Algorithm~\ref{alg:lll-sampler} for 2000 steps, we have:
\begin{equation}
T^{2000}=\begin{bNiceMatrix}[
  first-row,code-for-first-row=\normalsize,
  first-col,code-for-first-col=\normalsize,
]
 &s_1 &s_2 & s_3  & s_4 & s_5 & s_6 & s_7 & s_8 \\
s_1& \mathbf{1} & 0 & 0 & 0 & 0 & 0 & 0 & 0 \\
 s_2& 0 & \mathbf{1} & 0 & 0 & 0& 0 & 0 & 0\\
s_3& 0 & 0 &  \mathbf{1}  & 0 & 0& 0 & 0 & 0\\
s_4& 0 & 0 & 0 & \mathbf{1} & 0 & 0 & 0 & 0\\
s_5& \frac{1}{3} & \frac{1}{3} & \frac{1}{3} & 0& 0 & 0 & 0 & 0 \\
s_6 & \frac{1}{6} & \frac{1}{2} & \frac{1}{6} & \frac{1}{6} & 0 & 0 & 0 & \frac{1}{4} \\
s_7 & \frac{1}{3} & 0 & \frac{1}{3} & \frac{1}{3}& 0 & 0 &  0  & 0 \\
s_8& \frac{1}{6} & \frac{1}{6}  & \frac{1}{6}  & \frac{1}{2} & 0 & 0 &  0  & 0 \\
\end{bNiceMatrix},\quad xT^{2000}=\begin{bNiceMatrix}[
  first-row,code-for-first-row=\normalsize,
  first-col,code-for-first-col=\normalsize,
]
 &s_1 &s_2 & s_3  & s_4 & s_5 & s_6 & s_7 & s_8 \\
 &\frac{1}{4} & \frac{1}{4} & \frac{1}{4} & \frac{1}{4} & 0 & 0 & 0 & 0 \\
\end{bNiceMatrix}
\end{equation}
This implies Algorithm~\ref{alg:lll-sampler} outputs every valid assignment with the same probability in the uniform setting, which follows the result in Theorem~\ref{th:product-dist}.

\section{Running Time Analysis of Algorithm~\ref{alg:lll-sampler}} \label{appendix:time-analysis}
We dedicate this section to showing the running time of Algorithm~\ref{alg:lll-sampler} on a general weighted case.
The expected running time of Algorithm~\ref{alg:lll-sampler} is determined by the number of rounds of re-sampling.  
Algorithm~\ref{alg:lll-sampler} re-sample all the related random variables simultaneously in every single round.
However, it is hard to get an estimation of the exact total running time over the \textit{random variables}. Instead, we can only have a loose upper bound of the expected running time over the \textit{sequence of sampling record} (the sequence of violated constraints).

The overall structure of the proof is similar to the proof in~\citet[Theorem 13]{DBLP:journals/jacm/GuoJL19}. We show the difference in our proof at the end of this section.

\subsection{Definitions and Notations}
We define the following terms to simplify our notations.
\begin{definition}  \label{def:p_q_st}
Let $S_t$ be a subset of vertices in a dependency graph.
1) Define $p_{S_t}$ as the probability of  constraints in $S_t$ being violated:
\begin{equation} \label{eq:def-pst}
p_{S_t}=\p\left(\bigwedge_{c_i \in S_t}\neg{c}_i \right)
\end{equation}
{where we use $\neg{c}_i $ to indicate the constraint $c_i$ is violated.}
2) Define $q_{S_t}$ as the probability that only the constraints in $S_t$ are violated and nothing else. 
\begin{equation} \label{eq:definition-qst}
q_{S_t}=\p\left(\bigwedge_{c_i \in S_t}\neg{c}_i \wedge \bigwedge_{c_j \in \mathcal{C}\backslash S_t}c_j\right)
\end{equation}
{where $\bigwedge_{c_i \in S_t}\neg{c}_i $ corresponds to only the constraints in $S_t$ are violated and $\bigwedge_{c_j \in \mathcal{C}\backslash S_t}c_j$ corresponds to all the rest constraints are satisfied.}
So $q_{\{c_i\}}$ is the probability that only constraint $c_i$ is broken and all the rest still  hold. Similarly, $q_{\emptyset}$ denotes the probability that all the constraints are satisfied.
\end{definition}

\begin{lemma} \label{lem:recursion_start}
Given Definition~\ref{def:p_q_st}, we can further expand $q_{S_t}$ under Condition~\ref{cond:extreme}:
\begin{equation*}
q_{S_t}= p_{S_t}\p \left( \wedge_{c_j \in \mathcal{C}\backslash \Gamma(S_t)}c_j\right)
\end{equation*}
\begin{proof}
We can split $q_{S_t}$ into the probability of two independent events:
\begin{align*} 
    q_{S_t}&=\p\left(\bigwedge_{c_i \in S_t}\neg{c}_i \wedge \bigwedge_{c_j \in \mathcal{C}\backslash S_t}c_j\right) &\text{ By definition of $q_{S_t}$ in Equation~\eqref{eq:definition-qst}}\\
    &=\p\left(\bigwedge_{c_i \in S_t}\neg{c}_i \wedge \bigwedge_{c_j \in \mathcal{C}\backslash \Gamma(S_t)}c_j\right) \\
    &=\p\left(\bigwedge_{c_i \in S_t}\neg{c}_i\right) \p \left( \bigwedge_{c_n \in \mathcal{C}\backslash \Gamma(S_t)}c_j\right)\\
    &= p_{S_t}\p \left( \wedge_{c_j \in \mathcal{C}\backslash \Gamma(S_t)}c_j\right). &\text{ By definition of $p_{S_t}$ in Equation~\eqref{eq:def-pst}}
\end{align*}
The second equality holds because under  Condition~\ref{cond:extreme}, adjacent vertices have zero probability. In other words, when we observe that constraints in $S_t$ are violated, constraints in $\Gamma(S_t)\backslash S_t$ cannot be violated. The third equality holds because the random variables in $\var(S_t)$ are \textit{independent to} 
those variables in $\var(\mathcal{C}\backslash\Gamma(S_t))$. {So we can apply $P(AB)=P(A)P(B)$ when the events $A,B$  are independent to each other.}
\end{proof}
\end{lemma}

\begin{remark}[Equivalence of Record] \label{rem:equal}
At round $t$ of Algorithm~\ref{alg:lll-sampler},  it finds all the constraints $S_t$ that are broken ($\bigwedge_{c_i \in S_t}\neg{c}_i $), which implies the rest of the constraints $\mathcal{C}\backslash \Gamma(S_t)$ are satisfied ($\bigwedge_{c_j \in \mathcal{C}\backslash S_t}c_j$). Thus the probability of observing $S_t$ in the record is equivalent to the following:
\begin{align} \label{eq:st-expand}
\p(S_t)=\p\left(\bigwedge_{c_i \in S_t}\neg{c}_i \wedge \bigwedge_{c_j \in \mathcal{C}\backslash S_t}c_j\right)
\end{align}
\end{remark}

\begin{lemma} \label{eq:pair-prob}
Given a possible sampling record $S_1$ $\ldots$ $S_{t-1}$,$S_t$ by Algorithm~\ref{alg:lll-sampler}, the following equality holds for the pair $(S_{t-1},S_t)$:
\begin{equation*}
\sum_{S_t}q_{S_t}=\p(\wedge_{c_i \in \mathcal{C}\backslash \Gamma(S_{t-1})}c_i)
\end{equation*}
\begin{proof}
By Definition~\ref{def:record} of the sampling record, we have $S_t \subset \Gamma(S_{t-1})$. The  relationship of its complement would be:
\begin{equation*}
\mathcal{C}\backslash\Gamma(S_{t-1})\subset \mathcal{C}\backslash S_t.
\end{equation*}
Using the above result, we have:
\begin{equation}\label{eq:subset}
\p\left(\bigwedge_{c_j \in \mathcal{C}\backslash S_t}c_j \wedge \bigwedge_{c_k \in \mathcal{C}\backslash \Gamma(S_{t-1})}c_k \right)=
\p\left(\bigwedge_{c_j \in \mathcal{C}\backslash S_t}c_j \right)
\end{equation}

Based on Remark~\ref{rem:equal} and Baye's theorem, we have:
\begin{equation} \label{eq:bayes}
\begin{aligned}
\p(S_t |\wedge_{c_i \in \mathcal{C}\backslash \Gamma(S_{t-1})}c_i)&=\p\left(\bigwedge_{c_i \in S_t}\neg{c}_i \wedge \bigwedge_{c_j \in \mathcal{C}\backslash S_t}c_j \Big| \wedge_{c_k \in \mathcal{C}\backslash \Gamma(S_{t-1})}c_k \right) & \text{ By Equation~\eqref{eq:st-expand}}\\
&=\frac{\p\left(\bigwedge_{c_i \in S_t}\neg{c}_i \wedge \bigwedge_{c_j \in \mathcal{C}\backslash S_t}c_j \wedge \bigwedge_{c_k \in \mathcal{C}\backslash \Gamma(S_{t-1})}c_k \right)}{\p \left(\wedge_{c_i \in \mathcal{C}\backslash \Gamma(S_{t-1})}c_i\right) } &\text{By Bayes's formula}\\
&=\frac{\p\left(\bigwedge_{c_i \in S_t}\neg{c}_i \wedge \bigwedge_{c_k \in \mathcal{C}\backslash S_{t}}c_k \right)}{\p \left(\wedge_{c_i \in \mathcal{C}\backslash \Gamma(S_{t-1})}c_i\right) } &\text{ By Equation~\eqref{eq:subset}}\\
&= \frac{q_{S_t} }{\p \left(\wedge_{c_i \in \mathcal{C}\backslash \Gamma(S_{t-1})}c_i\right) }. &\text{ By definition of $p_{S_t}$ in Equation~\eqref{eq:definition-qst}}
\end{aligned}
\end{equation}

Since LHS of Equation~\eqref{eq:bayes} sums over all possible $S_t$ is one: $\sum_{S_t}\p(S_t |\wedge_{c_i \in \mathcal{C}\backslash \Gamma(S_{t-1})}c_i)=1$. Thus, summing over $S_t$ for the RHS of Equation~\eqref{eq:bayes}, we have:
\begin{equation} \label{eq:summation-st} 
\begin{aligned}
1=\sum_{S_t}\frac{q_{S_t} }{\p \left(\wedge_{c_i \in \mathcal{C}\backslash \Gamma(S_{t-1})}c_i\right) }=\frac{\sum_{S_t}q_{S_t} }{\p \left(\wedge_{c_i \in \mathcal{C}\backslash \Gamma(S_{t-1})}c_i\right) }
\end{aligned}
\end{equation}
In the second equality, the reason that we can move the summation operator to  the numerator is that the denominator is a constant \textit{w.r.t.} all possible $S_t$. To be specific, given $S_t\subseteq\Gamma(S_{t-1})$, we have  $S_t$ is independent to $\mathcal{C}\backslash \Gamma(S_{t-1})$.
Based on Equation~\eqref{eq:summation-st}, we finally obtain:
\begin{equation*}
\sum_{S_t}q_{S_t}=\p(\wedge_{c_i \in \mathcal{C}\backslash \Gamma(S_{t-1})}c_i).
\end{equation*}
The proof  is finished. 
\end{proof}
\end{lemma}

\begin{lemma} \label{lem:sample-record}
The probability of observing the sampling record  $S_1,\ldots,S_T$ by Algorithm~\ref{alg:lll-sampler} under Condition~\ref{cond:extreme} is:
\begin{equation}\label{pS_final}
    \p(S_1,\ldots ,S_T)=q_{S_T} \prod_{t=1}^{T-1} p_{S_t} 
\end{equation}
\begin{proof}
Given sampling record  $S_1,\ldots,S_{t-1}$, the conditional probability of observing the next record {$S_t,S'_t$} can be expanded  based on Lemma~\ref{lem:ratio-prob},
\begin{equation*}
\frac{\p(S_t | S_1, \ldots, S_{t-1})}{\p(S'_t | S_1, \ldots, S_{t-1})}=\frac{\p(S_t |\wedge_{c_i \in \mathcal{C}\backslash \Gamma(S_{t-1})}c_i)}{\p(S'_t |\wedge_{c_i \in \mathcal{C}\backslash \Gamma(S_{t-1})}c_i)}
\end{equation*}
Based on Equation~\eqref{eq:bayes}, we {can simplify the RHS of the above ratio equality and}] obtain:
\begin{equation*}
\frac{\p(S_t | S_1, \ldots, S_{t-1})}{\p(S'_t | S_1, \ldots, S_{t-1})}=\frac{q_{S_t}}{q_{S'_t}}
\end{equation*}
Because of $\sum_{S_t}\p(S_t | S_1, \ldots, S_{t-1})=1$ and Equation~\eqref{eq:summation-st}, we can get:
\begin{equation}\label{recursion_base}
  \p(S_t | S_1, \ldots, S_{t-1})=  \frac{q_{S_t}}{\p(\wedge_{c_i \in \mathcal{C}\backslash \Gamma(S_{t-1})}c_i)}
\end{equation}
We finally compute the probability of observing the sampling record $S_1\,\ldots, S_T$ by:
\begin{equation*}
\begin{aligned} 
    \p(S_1\,\ldots ,S_T)=&   \p(S_1)\prod_{t=2}^T\p(S_t | S_1,\ldots,S_{t-1})  &\text{ By Chain rule}\\
    =&q_{S_1}\prod_{t=2}^T \frac{q_{S_t}}{\p(\wedge_{c_i \in \mathcal{C}\backslash \Gamma(S_{t-1})}c_i)}  &\text{ By Equation~\eqref{recursion_base}}\\
    =&q_{S_T} \prod_{t=2}^T \frac{q_{S_{t-1}}}{\p(\wedge_{c_i \in \mathcal{C}\backslash \Gamma(S_{t-1})}c_i)}  &\text{ Left shift the numerator from $S_t$ to $S_{t-1}$}\\
    =&q_{S_T} \prod_{t=1}^{T-1} p_{S_t} & \text{Plugin  Lemma~\ref{lem:recursion_start}}
\end{aligned}    
\end{equation*}
The proof is finished.
\end{proof}
\end{lemma}

\subsection{An Upper Bound on Expected Running Time}
Suppose the expected number of samplings of constraints $c_i$ is $\e(T_i)$, then the total running time will be:
\begin{equation*}
\e(T) \le \sum_{i=1}^n \e (T_i)
\end{equation*}

Since each random variable has equal status, then the question comes down to the computation of individual $T_i$'s expectation. Let $S_1,\ldots,S_T$ be any record of the algorithm that successfully terminates, and $T_i(S_1,\ldots,S_T)$ be the total number of sampling related to constraint $c_i$ throughout this record.  Based on Lemma~\ref{lem:sample-record}, we have:
\begin{align*} \label{ti_formula}
    \e(T_i) &= \sum_{S_1,\ldots,S_T} \p(S_1,\ldots,S_T)T_i(S_1,\ldots,S_T) 
\end{align*}
{By far, we have shown the original proof of our work. We leave the difference between our proof with the existing one in Appendix~\ref{apx:difference}.}

The rest of the computation can be done in the same way as the proof in \citet{DBLP:journals/jacm/GuoJL19}. Thus we cite the necessary intermediate steps in the existing work and finish the proof logic for the coherence of the whole running time analysis.
\begin{lemma*}[\citet{DBLP:journals/jacm/GuoJL19} Lemma 12] \label{thm:runtime} Let $q_{\emptyset}$ be a non-zero probability of all the constraints are satisfied. Let $q_{\{c_j\}}$ denote the probability that only constraint $c_j$ is broken and the rest all hold. If $q_\emptyset>0$, then
$\e(T_i) ={q_{\{c_j\}}}/{q_\emptyset}$.
\end{lemma*}

After incorporating our fix, we can conclude the upper bound on the expected running time in Theorem~\ref{thm:time}.
\begin{theorem}[\citet{DBLP:journals/jacm/GuoJL19} Theorem 13] \label{thm:time} Under Condition~\ref{cond:extreme},
 the total number of re-samplings throughout the algorithm is then $\frac{1}{q_{\emptyset}}\sum_{j=1}^L q_{\{c_j\}}$.
\end{theorem}

\subsection{Difference to the Existing Proof}  \label{apx:difference}
 The main difference in the above proof to the existing proof in~\cite[Theorem 13]{DBLP:journals/jacm/GuoJL19} is that: based on Equation~\eqref{recursion_base} and~\eqref{eq:bayes}, we show
\begin{equation*}
\p(S_t | S_1, \ldots, S_{t-1})=\p(S_t |\wedge_{c_i \in \mathcal{C}\backslash \Gamma(S_{t-1})}c_i)
\end{equation*}
In \citet{DBLP:journals/jacm/GuoJL19}'s Equation~(9), the first step cannot holds without the above equality. The original paper uses this result directly without providing enough justification.

\section{Constrained MRF Model}\label{appendix:constrain-mrf}

\subsection{Single Variable Form of Constrained MRF}  \label{appendix:svf}
Here we provide an example of transforming MRF with pairwise and single potential functions into a single potential form by introducing extra variables. Given random variables $X_1,X_2,X_3$, we have the following example MRF model:
\begin{align*}
\phi_{\theta}(x_1,x_2,x_3)&=\theta_1 x_1+\theta_2 x_2+\theta_3 x_1 x_2\\
P_{\theta}(x)&=\frac{\exp(\phi_{\theta}(x_1,x_2,x_3))}{Z({\theta})}
\end{align*}

In the above formula, we have a cross term $x_1 x_2$. Two Boolean variables can have 4 different assignments in total. Therefore we can construct 4 extra Boolean variables to encode all these assignments. To illustrate, we introduce extra random variables $\hat{X}_{00}$, $\hat{X}_{01}$, $\hat{X}_{10}$, $\hat{X}_{11}$. We further introduce extra constraints:
 When $X_1=0,X_2=0$, the extra variable must take values: $\hat{X}_{00}=1$, $\hat{X}_{01}=0$, $\hat{X}_{10}=0$, $\hat{X}_{11}=0$. See the rest constraints in Table~\ref{tab:pairwise-to-single}.

\begin{table}[!ht]
    \centering
    \begin{tabular}{cc|cccc}
    \hline
        $X_1$ & $X_2$ &$\hat{X}_{00}$, $\hat{X}_{01}$, $\hat{X}_{10}$, $\hat{X}_{11}$  \\ \hline
        $0$ & $0$ & $1, 0, 0, 0$\\ 
        $0$ & $1$ & $0, 1, 0, 0$\\ 
        $1$ & $0$ & $0, 0, 1, 0$\\ 
        $1$ & $1$ & $0, 0, 0, 1$\\ 
    \hline
    \end{tabular}
    \caption{4 constraints for converting pairwise terms in the potential function into single variable form.}
    \label{tab:pairwise-to-single}
\end{table}
{Then the new potential function, including extended variables and pairwise to single variable constraints $\mathcal{C}$, is reformulated as:}
\begin{align*}
\hat{\phi}_{\theta}(x_1,x_2,x_3,\hat{x}_{00},\hat{x}_{01},\hat{x}_{10},\hat{x}_{11})&=\theta_1 x_1+\theta_2 x_2+\theta_3 \hat{x}_{00}+ \theta_3 \hat{x}_{01}+\theta_3 \hat{x}_{10}+\theta_3 \hat{x}_{11}\\
P_{\theta}(x|\mathcal{C})&=\frac{\exp(\hat{\phi}_{\theta}(x_1,x_2,x_3,\hat{x}_{00},\hat{x}_{01},\hat{x}_{10},\hat{x}_{11}))}{Z_{\mathcal{C}}(\theta)}
\end{align*}
For clarity, the newly added constraints do not impact Condition~\ref{cond:extreme}. Since the single variable transformation in the MRFs model originates from~\citet{Sang2005}, thus is not considered as our contribution.

\subsection{Gradient of $\log$-Partition Function $\nabla\log Z_\mathcal{C}(\theta)$}\label{appendix:partition}
We use the Chain rule of the gradient to give a detailed deduction of Equation~\eqref{eq:gradient}.
\begin{equation}\label{eq:full-gradient}
\begin{aligned}
\nabla\log Z_\mathcal{C}(\theta)=\frac{\nabla Z_\mathcal{C}(\theta)}{Z_\mathcal{C}(\theta)}=\frac{1}{Z_\mathcal{C}(\theta)}\nabla\sum_{x\in \mathcal{X}}\exp\left( \phi_{\theta}(x)\right)C(x)=\sum_{x\in\mathcal{X}} \frac{\exp(\phi_{\theta}(x))C(x) }{Z_\mathcal{C}(\theta)}{\nabla\phi_\theta(x)} =&\sum_{x\in \mathcal{X}} P_\theta(x|\mathcal{C}) \nabla\phi_{\theta}(x) \\
=&\mathbb{E}_{x\sim P_{\theta}(\tilde{x}|\mathcal{C})} \left({\nabla\phi_{\theta}(x)}\right)
\end{aligned}
\end{equation}
{The above result shows the gradient of the constrained partition function is equivalent to the expectation of the gradient of the potential function $\nabla\phi_{\theta}$ over the model's distribution (\textit{i.e.}, $P_{\theta}(\tilde{x}|\mathcal{C})$). Therefore, we transform the gradient estimation problem into the problem of sampling from the current MRF model.}

\section{Experiment Settings and Configurations}
\label{appendix:exp-set}
\subsection{Implementation Details} \label{appendix:implementation}

\subsubsection{Implementation of \nls} The proposed sampler can be implemented with Numpy, Pytorch, or Jax. We further offer a ``batch version'' implementation, that draws a batch of samples in parallel on GPU. The batched sampler is useful for those tasks that require a huge number of samples to estimate the gradient with a small approximation error. 

{In Section~\ref{sec:nls}, we define  vector of assignment  $x^t=(x^t_1, \dots, x^t_n)$,
where $x_i^t$ is the assignment of variable $X_i$ in the $t$-th round of Algorithm~\ref{alg:lll-sampler}. $x^t_i=1$ denotes  variable $X_i$ takes value $1$ (or true). In the batch version, we define the matrix for a batch  of assignments. Let $b$ be the batch size, we have}
\begin{equation*}
x^{t}=\begin{bmatrix}
x^t_{11} & \dots & x^t_{1n} \\
  \vdots & \ddots & \vdots \\
x^t_{b1} & \dots & x^t_{bn} \\
\end{bmatrix}
\end{equation*}
{In the following part, we provide the detailed computation pipeline for the batch version of the proposed algorithm.}

\subsubsection{Initialization} {The first step is to sample an initial assignment of $X$ from the given the marginal probability vector $P$:}
\begin{equation}
 x^1_{li} = \begin{cases}
 1&\text{if } u_{li}> P_i,\\
 0 &\text{otherwise}. \\
 \end{cases},\quad \text{ if} 1\le i\le n, 1\le l\le b
\end{equation}
{Here $u_{li}$ is sampled from the uniform distribution over $[0,1]$.}

\subsubsection{Check Constraint Satisfaction} { The second step extract which constraint is violated. Given an assignment $x^t$ at round $t\ge 1$, tensor $W$ and matrix $b$, the computation of tensor $Z^t$ is:}
\begin{equation*}
Z^t_{lik} =\sum_{i=1}^n W_{ikj}  x_{lj}^t+b_{lik},
\end{equation*}
{The special  multiplication between tensor and matrix can be efficiently implemented with the  Einstein summation\footnote{\url{https://github.com/dgasmith/opt_einsum}}.
Note that $Z^t_{ljk}=1$ indicates for $l$-th batched assignment $x_l$, the $k$-th literal of $j$-th clause is true (takes value $1$). Next, we compute $S^t_{lj}$ as:}
\begin{align*}
S^t_{lj} &=1-\max_{1\le k\le K} Z_{ljk}, \quad \text{ for } 1\le j\le L, 1\le l\le b
\end{align*}
{Here $S^t_{lj}=1$ indicates $x_l^t$ violates $j$-th clause.
We can check $\sum_{l=1}^b\sum_{j=1}^LS^t_{lj}\neq 0$ to see if any clause is violated for the current batch of assignments, which corresponds to $\sum_{i=1}^bC(x_l)= 0$. }

\subsubsection{Extract Variables in Violated Clauses} { We extract all the  variables that require resampling based on vector $S^t$ computed from the last step.
The vector of the resampling indicator matrix $A^t$ can be computed as:}
\begin{equation*}
A^t_{li}=\mathbf{1}\left(\sum_{j=1}^L{S_{lj}^t} V_{ji}\ge 1\right),\quad \text{ for } 1\le i\le n, 1\le l\le b
\end{equation*}
where $\sum_{j=1}^L{S_{lj}^t} V_{ji}\ge 1$ implies $X_{li}$ requires resampling.

\subsubsection{Resample} {Given the marginal probability vector $P$, resample indicator matrix $A^t$ and assignment matrix $x^t$, we draw a new random sample $x^{t+1}$. }
\begin{equation*}
 x_{li}^{t+1}=\begin{cases}
 (1-A^t_{li}) x_{li}^t+A^t_{li}  &\text{if } u_{{li}}>P_i,\\
 (1-A^t_{li}) x_{li}^t &\text{otherwise}.
 \end{cases}\quad \text{ for } 1\le i\le n, 1\le l\le b
\end{equation*}
{where $u_{li}$ is drawn from the uniform distribution in $[0, 1]$. }

{Since GPUs are more efficient at computing tensor, matrix, and vector operations but are slow at processing for loops. Drawing a batch of samples using the above extended computational pipeline is much faster than using a for loop over the computational pipeline in Section~\ref{sec:nls}.}

The sampler is involved with one hyper-parameter $T_{\textit{tryout}}$. \nls  would terminate when it reaches $T_{\textit{tryout}}$ number of re-samples. This practice is commonly used to handle randomized programs that might run forever in rare cases. 

\begin{figure*}[!t]
\centering
\includegraphics[width=0.99\linewidth]{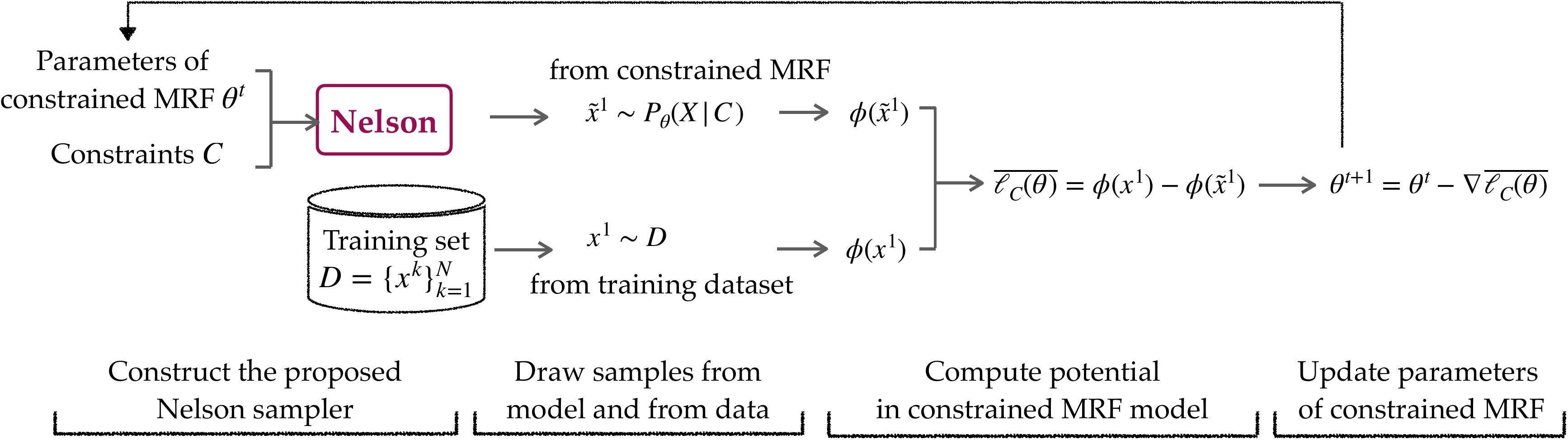}
\captionof{figure}{Implementation pipeline of the \nls-CD algorithm with $m=1$. The proposed \nls can be efficiently adapted to a Pytorch-based machine learning library and enforces constraint satisfaction during learning.}\label{fig:pipeline}
\end{figure*}

\subsubsection{Implementation of  Algorithm~\ref{alg:main}}  We first use the Constraints $\mathcal{C}$ and parameters $\theta^t$ to build the current \nls module.  Then we draw $m$ samples from \nls module $\{\tilde{x}^j\}_{j=1}^m$ and draw from dataset randomly $m$ samples $\{x^j\}_{j=1}^m$. Continuing from that point, we compute the potential value from the two sets of inputs, i.e., $\{ \phi_{\theta}(\tilde{x}^j)\}_{j=1}^m$ and $\{ \phi_{\theta}(\tilde{x}^j)\}_{j=1}^m$. Pytorch would be slow if we compute each potential's gradient using a for-loop. To bypass this problem, we instead compute the following:
\begin{equation} \label{eq:emp-gradient}
\overline{\ell_{\mathcal{C}}(\theta)}= \frac{1}{m}\sum_{j=1}^m \phi_{\theta}(x^j)-\frac{1}{m}\sum_{j=1}^m \phi_{\theta}(\tilde{x}^j).
\end{equation}
Following that, we call the PyTorch library's gradient function, which computes exactly
\begin{equation*}
\begin{aligned}
\nabla\overline{\ell_{\mathcal{C}}(\theta)}&=\nabla\left(\frac{1}{m}\sum_{j=1}^m \phi_{\theta}(x^j)-\frac{1}{m}\sum_{j=1}^m \phi_{\theta}(\tilde{x}^j)\right)=\frac{1}{m}\sum_{j=1}^m \nabla\phi_{\theta}(x^j)-\frac{1}{m}\sum_{j=1}^m \nabla\phi_{\theta}(\tilde{x}^j)
\end{aligned}
\end{equation*}
Note that $\nabla\overline{\ell_{\mathcal{C}}(\theta)}$ recovers the result in Equation~\eqref{eq:gradient}. Finally, we update the parameters $\theta$. The proposed \nls module and the neural network are computed on the same GPU device. This allows us to exploit the parallel computing power of modern GPUs and remove time for the data transfer from CPU to GPU or vice versa.   See Figure~\ref{fig:pipeline} for a visualized overview of the implementation with Pytroch.

\subsection{Learn Random K-SAT Solutions with Preference}
\subsubsection{Task Definition}  We are given a training set $\mathcal{D}$ containing some preferred assignments $\mathcal{D}=\{x^j\}_{j=1}^N$ for the corresponding CNF formula $c_1\wedge\ldots\wedge  c_L$. We require the CNF formula to be true. This means, by the definition of CNF formulas, that every clause has to be satisfied. These clauses become our set of constraints. Under the constrained MRF model, the learning task is to maximize the log-likelihood of the assignments seen in the training set $\mathcal{D}$. The inference task is to generate valid solutions from  the learned model's distribution~\cite{DBLP:conf/aiia/DodaroP19,DBLP:conf/sac/RosaGO11}.


\subsubsection{Dataset} We denote the Boolean variables' size in $K$-SAT as the ``problem size''. We consider several datasets of different problem sizes generated from CNFGen\footnote{\url{https://github.com/MassimoLauria/cnfgen}}~\cite{DBLP:conf/sat/LauriaENV17} random $K$-SAT functions. $K$ is fixed as $5$; the number of variables and clauses are kept the same,  ranging from $10$ to $1500$. We generate $100$ different CNF formulas  for every problem size. To generate the training set $\mathcal{D}$, we use the Glucose4 solver from PySAT\footnote{\url{https://pysathq.github.io/}} library~\cite{DBLP:conf/sat/IgnatievMM18} to generate $200$ assignments randomly as the preferred assignments for every formula. 

It should be pointed out that we don't consider datasets like SATLIB and SAT competitions. It is mainly because these datasets are hard instances with a much larger input space but a limited number of solutions. \nls would generally take exponential time to find these solutions, just like finding needles in a haystack. The other reasons are that using neural networks to learn these limited assignments is straightforward since we can simply hard-wire the network to memorize all the valid assignments. The main purpose of this work is to let a constrained MRF learn a representation for the underlying preference pattern, not create a neural solver that can generate valid assignments for any CNF formula. Thus, we conform to the settings of the easy formula where obtaining valid solutions is easy.

\subsection{Learn Sink-Free Orientation in Undirected Graphs}
\textbf{Task Definition} In graph theory, a \textit{sink-free} orientation of an undirected graph is a choice of orientation for each edge such that every vertex has at least one outgoing edge~\cite{DBLP:journals/combinatorics/CohnPP02}. It has wide applications in robotics routing and IoT network configuration~\cite{takahashi2009communication}.  The Constraints for this problem are that every vertex has at least one outgoing edge after orientation. As stated in~\cite{DBLP:journals/jacm/GuoJL19}, these constraints satisfy Condition~\ref{cond:extreme}.

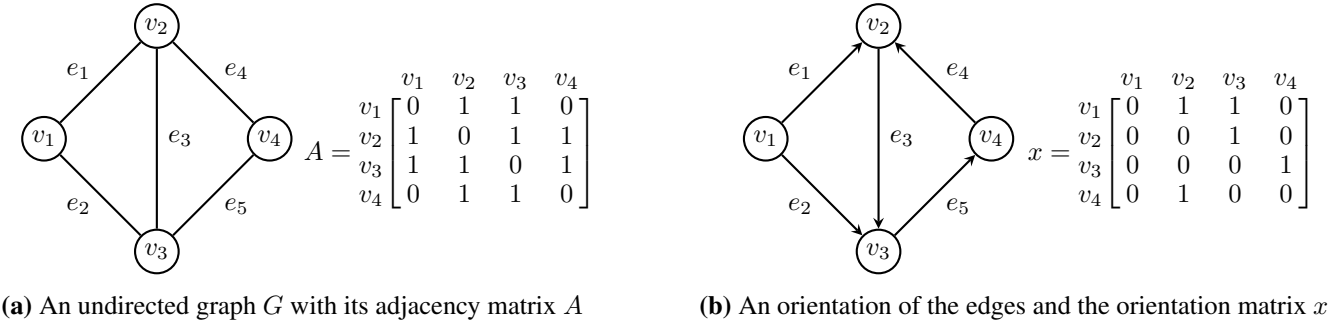
\begin{figure}[t]
\centering
\begin{tikzpicture}[xscale=0.15, yscale=0.15, inner sep=2pt, >=stealth]
    \draw (0,10)  node[circle,thick,draw] (v1) {$v_1$}; 
    \draw (10,20) node[circle,thick,draw] (v2) {$v_2$};
    \draw (10,0)  node[circle,thick,draw] (v3) {$v_3$}; 
    \draw (22,-5) node[thick] (v4) {\textbf{(a)} An undirected graph $G$ with its adjacency matrix $A$}; 
    \draw (20,10) node[circle,thick,draw] (v4) {$v_4$}; 
    \draw (36,10)  node (v5) {$A=\begin{bNiceMatrix}[
  first-row,code-for-first-row=\normalsize,
  first-col,code-for-first-col=\normalsize,
]
 &v_1 &v_2 & v_3  & v_4 \\
v_1& 0 & 1 & 1 & 0  \\
 v_2& 1 & 0 & 1 & 1\\
v_3& 1 & 1 &   0 & 1 \\
v_4& 0 & 1 & 1 & 0  \\
\end{bNiceMatrix}$}; 
    \draw[-,thick] (v1) -- (v2);
    \draw[-,thick] (v1) -- (v3);
    \draw[-,thick] (v2) -- (v3);
    \draw[-,thick] (v4) -- (v2); 
    \draw[-,thick] (v3) -- (v4);  
    \draw (4,16)  node[text width = 6mm] () {$e_1$}; 
    \draw (4,4)   node[text width = 6mm] () {$e_2$};
    \draw (13,10) node[text width = 6mm] () {$e_3$};
    \draw (18,16) node[text width = 6mm] () {$e_4$}; 
    \draw (18,4)  node[text width = 6mm] () {$e_5$};
  \end{tikzpicture}
 \hfill
\begin{tikzpicture}[xscale=0.15, yscale=0.15, inner sep=2pt, >=stealth]
    \draw (0,10)  node[circle,thick,draw] (v1) {$v_1$}; 
    \draw (10,20) node[circle,thick,draw] (v2) {$v_2$};
    \draw (10,0)  node[circle,thick,draw] (v3) {$v_3$}; 
    \draw (22,-5) node[thick] (v4) {\textbf{(b)} An orientation of the edges and the orientation matrix $x$}; 
    \draw (20,10) node[circle,thick,draw] (v4) {$v_4$}; 
     \draw (36,10)  node (v5) {$x=\begin{bNiceMatrix}[
  first-row,code-for-first-row=\normalsize,
  first-col,code-for-first-col=\normalsize,
]
 &v_1 &v_2 & v_3  & v_4 \\
v_1& 0 & 1 & 1 & 0  \\
 v_2& 0 & 0 & 1 & 0\\
v_3& 0 & 0 &   0 & 1 \\
v_4& 0 & 1 & 0 & 0  \\
\end{bNiceMatrix}$}; 
    \draw[->,thick] (v1) -- (v2);
    \draw[->,thick] (v1) -- (v3);
    \draw[->,thick] (v2) -- (v3);
    \draw[->,thick] (v4) -- (v2); 
    \draw[->,thick] (v3) -- (v4);  
    \draw (4,16)  node[text width = 6mm] () {$e_1$}; 
    \draw (4,4)   node[text width = 6mm] () {$e_2$};
    \draw (13,10) node[text width = 6mm] () {$e_3$};
    \draw (18,16) node[text width = 6mm] () {$e_4$}; 
    \draw (18,4)  node[text width = 6mm] () {$e_5$};
  \end{tikzpicture}
\caption{{\textbf{(a)} An un-directed graph $G(V,E)$ where the vertices are $V=\{v_1,v_2,v_3,v_4\}$ and the un-directed edges are $E=\{e_1=(v_1,v_2),e_2=(v_1,v_3),e_3=(v_2,v_3),e_4=(v_2,v_4),e_5=(v_3,v_4)\}$. \textbf{(b)} A possible sink-free orientation of the edges in the graph and its matrix representation $x$, where every vertex has at least one outgoing edge. }}
\label{fig:sinkfreeex}
\end{figure}

{See Figure~\ref{fig:sinkfreeex} for an example graph and one possible sink-free edge orientation. We define binary variables $X_1,\ldots, X_m$, and associate variable $X_i$ to edge $e_i$ for $1\le i\le m$. Variable $X_i$ takes value $1$ if the edge orientation is $v_i\to v_j$ where $i<j$. Otherwise, $X_i$ takes value $0$. The constraints are:}
\begin{equation*}
\mathcal{C}=(X_1\vee X_2)\wedge(\neg X_1\vee X_3\vee\neg X_4)\wedge(\neg X_2\vee\neg X_3\vee X_5)\wedge(X_4\vee \neg X_5)
\end{equation*}
{where the single constraint  $c_1=(X_1\vee X_2)$ corresponds to vertex $v_1$, constraint $c_2=(\neg X_1\vee X_3\vee\neg X_4)$ corresponds to vertex  $v_2$, constraint $c_3=(\neg X_2\vee\neg X_3\vee X_5)$  corresponds to vertex  $v_3$, and constraint $c_4=(X_4\vee \neg X_5)$ corresponds to vertex  $v_4$. The orientation assignment matrix $x$ shown in Figure~\ref{fig:sinkfreeex}(b) implies: $X_1=1,X_2=1,X_3=1,X_4=0,X_5=1$.}

\subsubsection{Notations} Let  graph $G(V,E)$ be an  un-directed graph; its
 adjacency matrix $A$ that represents graph connectivity is:
\begin{equation}\label{eq:adjacency}
A_{ij}=\begin{cases}
{1}&\text{If } (v_i,v_j)\in E\\
0&\text{otherwise}
\end{cases}
\end{equation}
A possible assignment for the orientation of every edge can be represented as a matrix $x\in\{0,1\}^{|V|\times |V|}$:
\begin{equation}\label{eq:orientation-assign}
x_{ij}=\begin{cases}
1& \text{if the edge orientation is } v_i\to v_j \\
0&\text{otherwise}
\end{cases}
\end{equation}
In the constrained MRF model defined in Eq.~\eqref{eq:constr_mrf_single}, the potential function of one orientation of all edges is
\begin{equation*}
 \phi_{\theta}(x)=\sum_{i=1}^{|V|}\sum_{j=1}^{|V|}\theta_{ij}A_{ij}x_{ij}
\end{equation*}
A single constraint for vertex $v_k$ is $c_k(x)=\mathbf{1}\left(\sum_{j=1}^nA_{k,j}x_{k,j}=1\right)$. If there is no ongoing edge of vertex $v_k$. The  constraint function $C(x)$ is defined as: $\prod_{i=1}^nc_k(x)$. In Algorithm~\ref{alg:lll-sampler} step 1, edge $(v_i,v_j)$ will pick the orientation $v_i\to v_j$  with probability:
\begin{equation*}
\frac{\exp(\theta_{ij}A_{ij}x_{ij})}{\exp(\theta_{ji}A_{ji}x_{ji})+\exp(\theta_{ij}A_{ij}x_{ij})}
\end{equation*}

\subsubsection{Dataset} {We use the NetworkX\footnote{https://networkx.org/} package to generate random Erdos Renyi graph with edge probability $0.55$. The problem size refers to the number of vertices in the graph, we range the problem size from 10 to 100. For each problem size, we generate 100 different random undirected graphs. We then convert the graph into CNF form using the above edge-variable conversion rule. Afterward, we follow the same processing steps as the previous problem that learn preferential solution distribution for random K-SAT. }.


\subsection{Learn Vehicle Delivery Routes}
Given a set of locations to visit, the task is to generate a sequence to visit these locations in which each location is visited once and only once and the sequence closely resembles the trend presented in the training data.  The training data are such routes collected in the past.  
The dataset is constructed from TSPLIB, which consists of $29$ cities in Bavaria, Germany. 
In Figure~\ref{fig:route}, we see  \nls can obtain samples of this delivery problem highly efficiently. 

A possible travel plan can be represented as a matrix $x\in\{0,1\}^{|V|\times |V|}$:
\begin{equation}\label{eq:route}
x_{ij}=\begin{cases}
1& \text{if edge } v_i\to v_j \text{ is selected}\\
0&\text{otherwise}
\end{cases}
\end{equation}
The constraints are that every  routing plan should visit every location once and only once.

Similarly, in the constrained MRF model defined in Eq.~\eqref{eq:constr_mrf_single}, the potential function of the vehicle routing plan is
\begin{equation*}
 \phi_{\theta}(x)=\sum_{i=1}^{|V|}\sum_{j=1}^{|V|}\theta_{ij}A_{ij}x_{ij}
\end{equation*}


\subsection{Detailed Baselines Configuation} 
In terms of sampling-based methods, we consider:
\begin{itemize}
\item Gibbs sampler~\cite{carter1994gibbs}, a special case of MCMC that is  widely used in training MRF models. In each step, the Gibbs algorithm samples 
one dimension based on a conditional marginal distribution. We follow this implementation\footnote{\url{https://github.com/Fading0924/BPChain-CD/blob/master/mrf.py}}.
    \item Weighted SAT samplers, including WAPS\footnote{\url{https://github.com/meelgroup/waps}}~\cite{DBLP:conf/tacas/GuptaSRM19}, WeightGen\footnote{\url{https://bitbucket.org/kuldeepmeel/weightgen/src/master/}}~\cite{DBLP:conf/aaai/ChakrabortyFMSV14} and XOR sampler\footnote{\url{https://cs.stanford.edu/~ermon/code/srcPAWS.zip}}~\cite{DBLP:conf/nips/ErmonGSS13,DBLP:conf/uai/DingX21}.
    \item Uniform SAT samplers, including UniGen\footnote{\url{https://github.com/meelgroup/unigen}}~\cite{DBLP:conf/cav/SoosGM20}, QuickSampler\footnote{\url{https://github.com/RafaelTupynamba/quicksampler}}~\cite{DBLP:conf/icse/DutraLBS18}, CMSGen\footnote{\url{https://github.com/meelgroup/cmsgen}}~\cite{DBLP:conf/fmcad/GoliaSCM21} and KUS\footnote{\url{https://github.com/meelgroup/KUS}}~\cite{DBLP:conf/lpar/SharmaGRM18}.
\end{itemize}
Currently, there are only GPU-based SAT solvers~\cite{DBLP:conf/sat/PrevotSM21,mahmoud2022gpu} and model counters~\cite{DBLP:conf/cp/FichteHZ19}, GPU-based SAT samplers are not available by far.

\subsection{Detailed Definition of Evaluation Metrics}
In terms of evaluation metrics, we consider
\begin{itemize}
    \item Training time per epoch. The average time for the whole learning method  to finish one epoch with each sampler.
    \item Validness. The learned model is adopted to generate assignments and we evaluate the percentage of generated assignments that satisfy the constraints.
    \item  Mean Averaged Precision (MAP$@10$).  This is a ranking-oriented metric that can evaluate the closeness of  the learned MRF  distribution to the  goal distribution. If the model learns the goal distribution in the training set, then it would assign a higher potential value to those assignments in the training set than all the rest unseen assignments. Based on this principle,
    we randomly pick two sets of inputs in those valid assignments: seen assignments from the training set  and unseen  assignments that are randomly generated. We use the value of factor potential $\phi(x)$ to rank those assignments in ascending order.  Next, we check how many preferred solutions can fall into the Top-$10$ by computing the following
\begin{equation*}
\text{MAP}@10=\sum_{k=1}^{10}\frac{\#\text{preferred assignments among top-} k}{k}
\end{equation*}

    \item $\log$-likelihood of assignments in the training set $\mathcal{D}$. The model that attains the highest $\log$-likelihood learns the closest  distribution to the training set.  Specifically, given a training set $\mathcal{D}=\{x^k\}_{k=1}^N$ and parameters $\theta$, the log-likelihood value is:
\begin{equation} 
\begin{aligned}
{\frac{1}{N}\sum_{k=1}^N\log P_{\theta}(X=x^k|\mathcal{C})}&=\frac{1}{N}\sum_{k=1}^N\phi_\theta(x^k)- \log Z_\mathcal{C}(\theta)
\end{aligned}
\end{equation}
We use the ACE algorithm to compute the approximated value of $\log Z_\mathcal{C}(\theta) $\footnote{\url{http://reasoning.cs.ucla.edu/ace/moreInformation.html}}.

    \item Approximation Error of $\nabla \log Z_{\mathcal{C}}(\theta)$, that is the $L_1$ distance between the exact gradient of $ \log Z_{\mathcal{C}}(\theta)$ in Eq.~\eqref{eq:gradient} and the empirical gradient from the sampler. For small problem sizes, we enumerate all $x\in \mathcal{X}$ to get the exact gradient and draw samples $\{\tilde{x}^j\}_{j=1}^m$ with $m=2000$  from every sampler for approximation.
    \begin{equation*}
    \Big|\underbrace{\sum_{x\in\mathcal{X}} \frac{\exp\left( \sum_{j=1}^n\theta_jx_j\right)C(x) }{Z_\mathcal{C}(\theta)}{x_i}}_{\text{{Exact gradient term}}} \ - \ \underbrace{\sum_{j=1}^m \tilde{x}^j_i}_{\text{{Estimated gradient with sampler}}}\Big|
    \end{equation*}
    For fixed parameter $\theta$, the best sampler would attain the smallest approximation error.
\end{itemize}

\subsection{Hyper-parameter Settings} 
 In the implementation of \nls, we set the maximum tryout of resampling as $T_{tryout}=1000$ for all the experiments and all the datasets.

For the hyper-parameters used in learning the constrained MRF, we set the number of samples from the model to be $m=200$, the learning rate $\eta$ is configured as $0.1$ and the total learning iterations are $T_{\max}=1000$.

\begin{figure*}[!ht]
    \centering
    \includegraphics[width=0.245\linewidth]{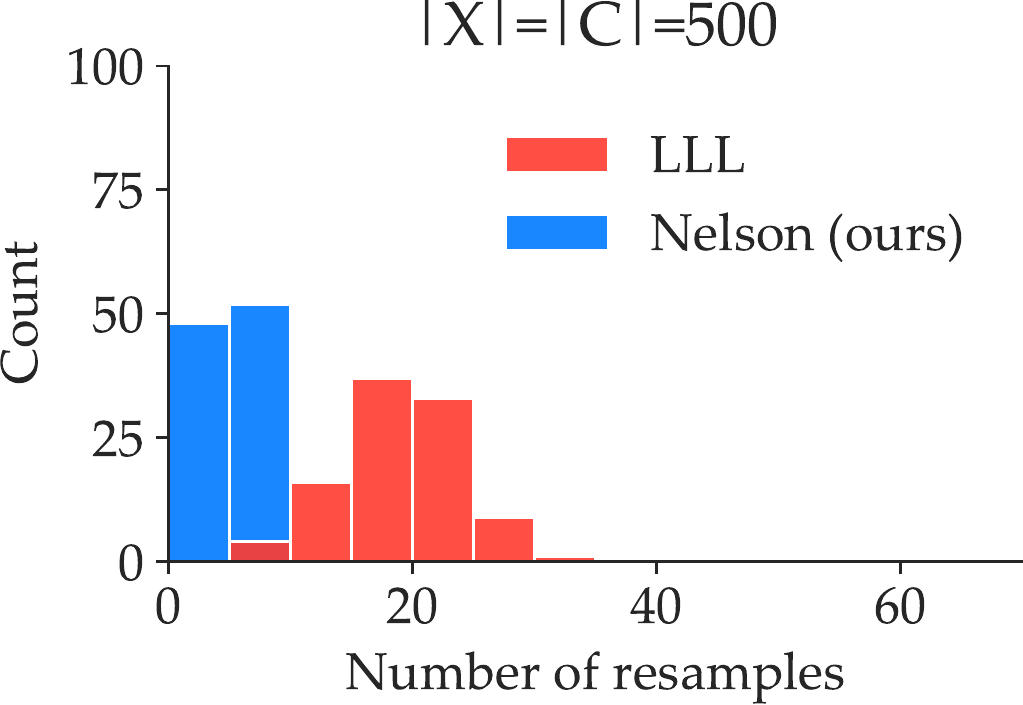}
    \includegraphics[width=0.245\linewidth]{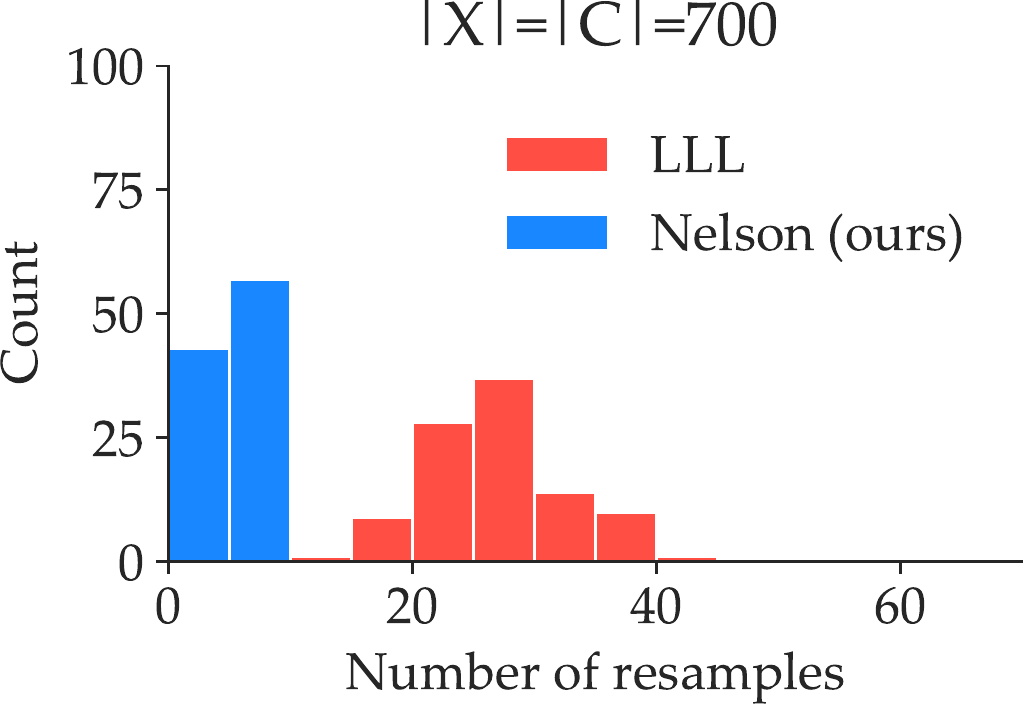}
    \includegraphics[width=0.245\linewidth]{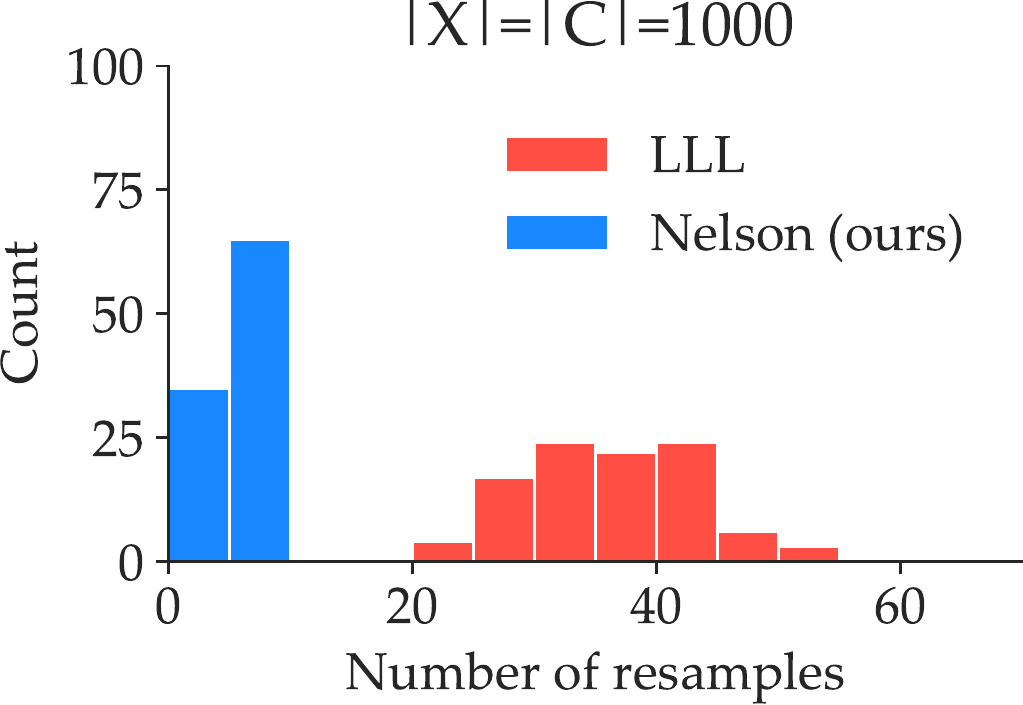}
    \includegraphics[width=0.245\linewidth]{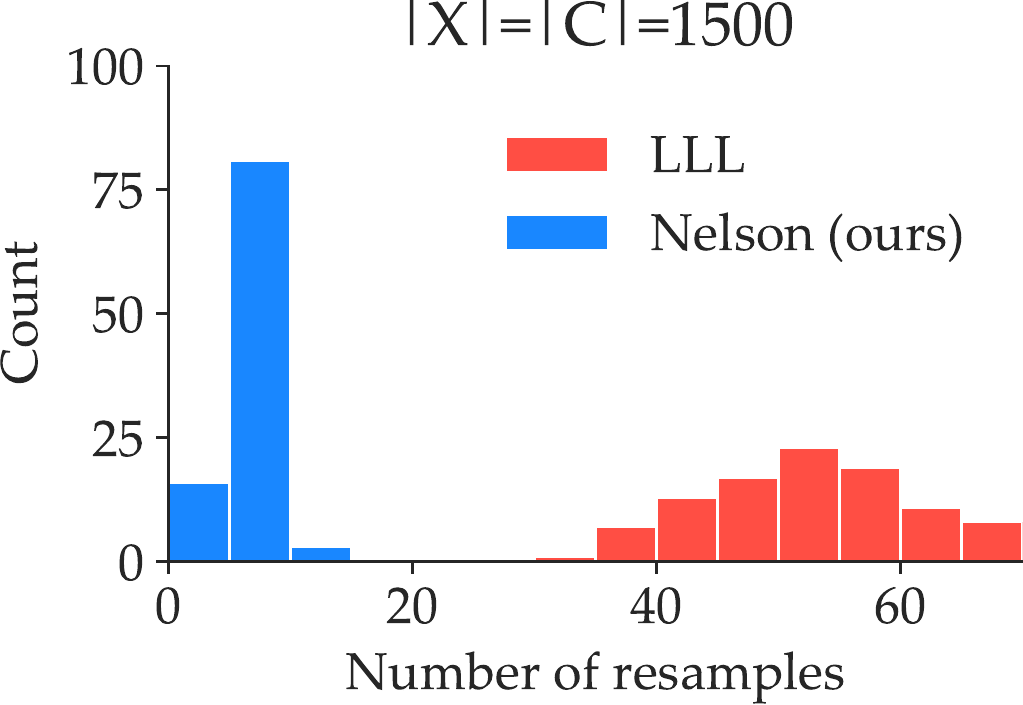}
    \begin{center}
      (a) Uniform Case.
    \end{center}
    \includegraphics[width=0.245\linewidth]{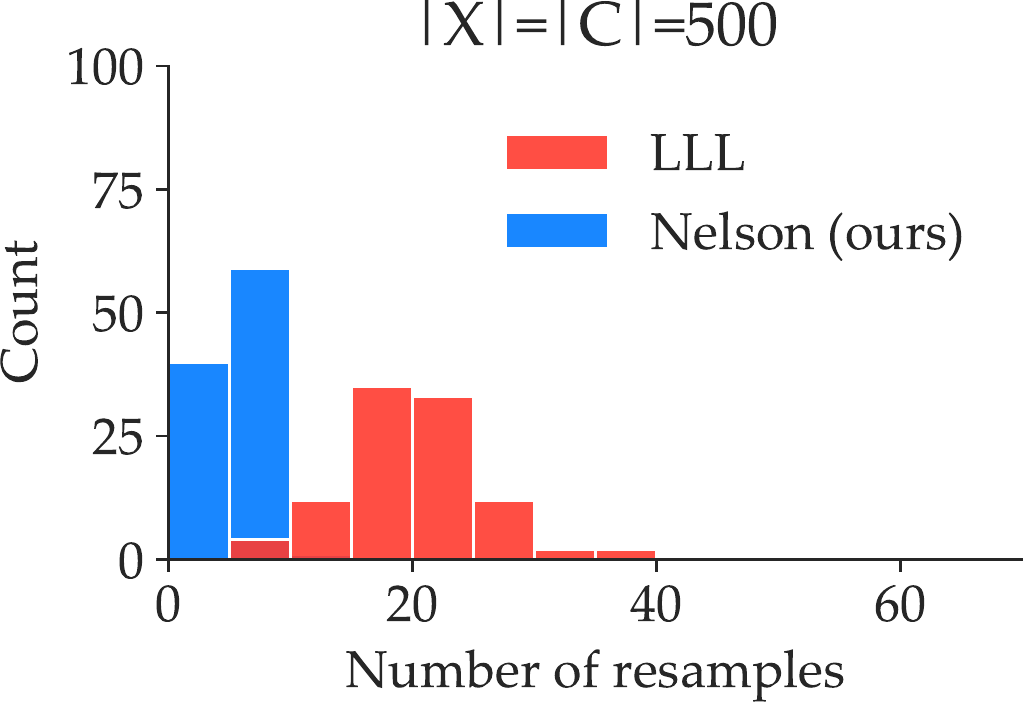}
    \includegraphics[width=0.245\linewidth]{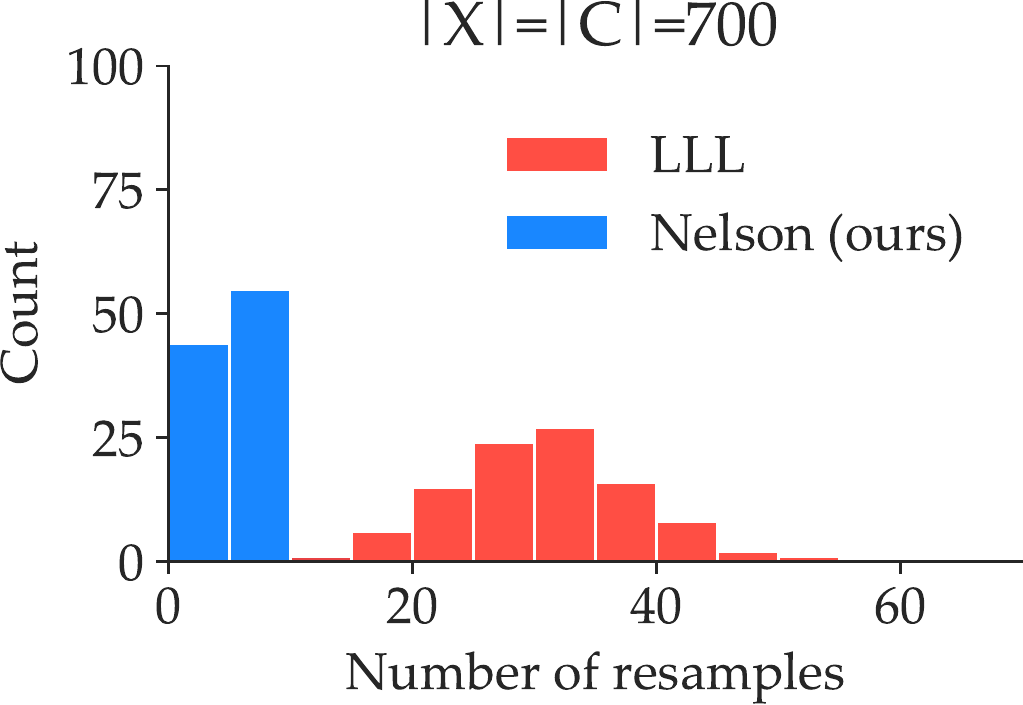}
    \includegraphics[width=0.245\linewidth]{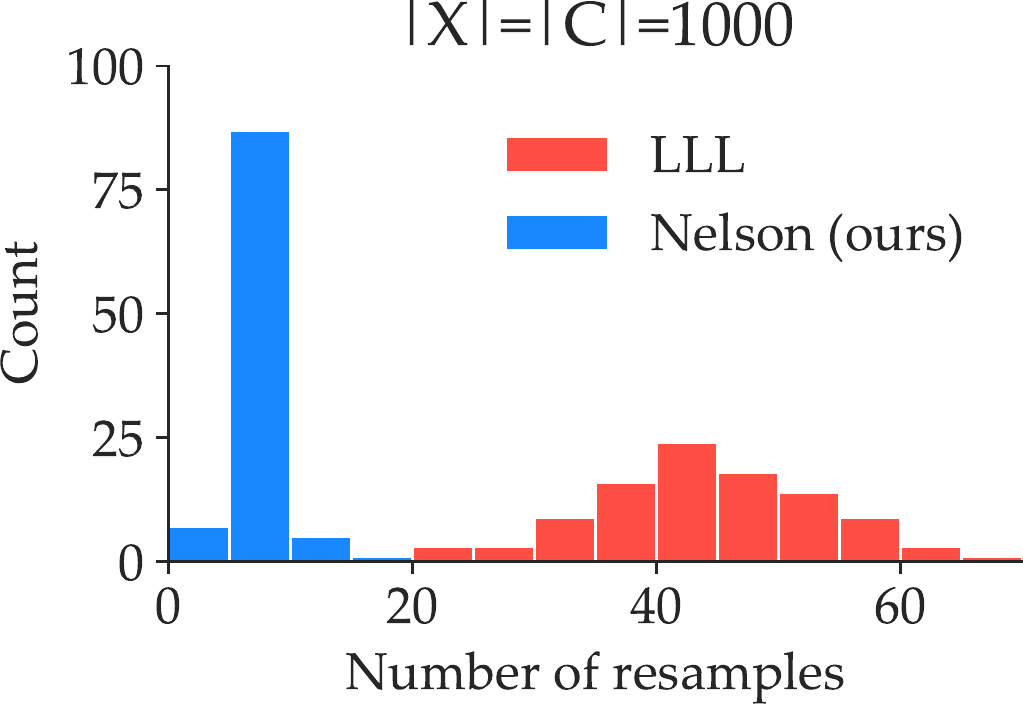}
    \includegraphics[width=0.245\linewidth]{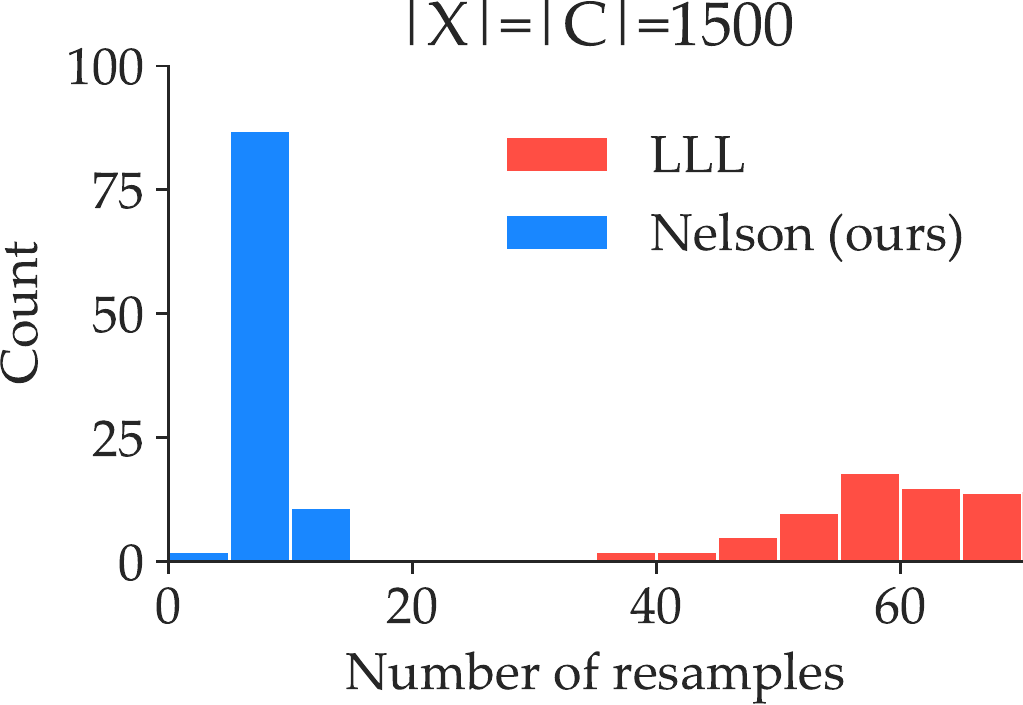}
    \begin{center}
      (b) Weighted Case.
    \end{center}
    \caption{The distribution of resampling steps in the \nls and Algorithmic-LLL~\cite{DBLP:journals/jacm/MoserT10}. Both of them get a valid sample within $T_{\mathit{tryouts}}$. \nls takes much fewer resamples than Algorithmic-LLL because it resamples all the violated clauses at every iteration while Algorithmic-LLL only resamples one of them.}\label{fig:ap:resamples}
\end{figure*}

\end{document}